\documentclass[letterpaper, 10pt, journal, final, twocolumn]{IEEEtran}
\usepackage{graphicx}
\usepackage{amsmath,amssymb,amsfonts}
\usepackage{xcolor}
\usepackage{mathptmx}
\usepackage[mathscr]{euscript}
\usepackage{amsthm}
\usepackage{cases}
\usepackage{setspace}
\usepackage{empheq}
\usepackage{caption}
\usepackage{notoccite}
\usepackage{cite}
\usepackage{algorithmic}
\usepackage{textcomp}
\usepackage{ulem}
\usepackage{flushend}
\usepackage{enumitem}
\usepackage{booktabs}
\usepackage{mdwtab,tabularx}
\usepackage{multirow}
\usepackage{multicol}
\usepackage{subfig}
\usepackage{caption}
\usepackage{array}

\newcommand{\tabincell}[2]{\begin{tabular}{@{}#1@{}}#2\end{tabular}}

\newcommand{\blue}{\color{blue}}

\begin{document}
\title{
{\Huge DARE: AI-based Diver Action Recognition System using Multi-{Channel} CNNs for AUV Supervision}\\ \vspace{6pt}
}

\author{\IEEEauthorblockN{Jing Yang\IEEEauthorrefmark{2}\IEEEauthorrefmark{1}, James P. Wilson\IEEEauthorrefmark{2} and Shalabh Gupta\IEEEauthorrefmark{2}
}
\\ \vspace{-18pt}
\thanks{\IEEEauthorrefmark{2} Department of Electrical and Computer Engineering, University of Connecticut, Storrs, CT.}
\thanks{\IEEEauthorrefmark{1} Corresponding author: Shalabh Gupta (email: shalabh.gupta@uconn.edu)}
}

\maketitle
\thispagestyle{empty}

\begin{abstract}
With the growth of sensing, control and robotic technologies, autonomous underwater vehicles (AUVs) have become useful assistants to human divers for performing various underwater operations. In the current practice, the divers are required to carry expensive, bulky, and waterproof keyboards or joystick-based controllers for supervision and control of AUVs. Therefore, diver action-based supervision is becoming increasingly popular because it is convenient, easier to use, faster, and cost effective. However, the various environmental, diver and sensing uncertainties present underwater makes it challenging to train a robust and reliable diver action recognition system. In this regard, this paper presents DARE, a diver action recognition system, that is trained based on Cognitive Autonomous Driving Buddy (CADDY) dataset, which is a rich set of data containing images of different diver gestures and poses in several different and realistic underwater environments. DARE is based on fusion of stereo-pairs of camera images using a multi-channel convolutional neural network supported with a systematically trained tree-topological deep neural network classifier to enhance the classification performance. DARE is fast and requires only a few milliseconds to classify one stereo-pair, thus making it suitable for real-time underwater implementation. DARE is comparatively evaluated against several existing classifier architectures and the results show that DARE supersedes the performance of all classifiers for diver action recognition in terms of overall as well as individual class accuracies and F1-scores.   
\end{abstract}
\vspace{-6pt}

\vspace{-6pt}
\section{Introduction}
\vspace{-0pt}
Underwater robots (e.g., autonomous underwater vehicles (AUVs)) have become vital assistants to human operators for a variety of tasks~\cite{Yuh2011_Applications_marinerobot} including search and exploration~\cite{song2018varepsilon,SG19,Katzschmann2018}, 3-D seafloor mapping\cite{Shen2017,Negahdaripour2003}, ocean resource analysis\cite{Wakita2010_AUVResources}, marine data collection\cite{Somers2016}, target tracking~\cite{HGW17,HGW2019,Shojaei2017}, ocean demining~\cite{MGR11,Acar2003}, oil spill cleaning \cite{SGH13,Kumar2020} and underwater structural inspection and repairing\cite{Foresti2001}. Since underwater environments can be hazardous, human divers can utilize robots for performing heavy or risky tasks,  such as lifting heavy parts \cite{Sivcev2018_Underwatermanipulators_Review}. On the other hand, underwater robots can also perform tasks which require precision and delicacy, such as gathering light-weight biological samples\cite{Galloway2016_Softroboticgrippers}. Therefore, it is becoming increasingly important to develop tools and methods that facilitate cost-effective, fast, and reliable communication between the divers and their robot counterparts for safe and rapid completion of underwater tasks.

\vspace{-3pt}
\subsection{Motivation} 
\begin{figure}[t] 
\centering
\includegraphics[width=0.45\textwidth]{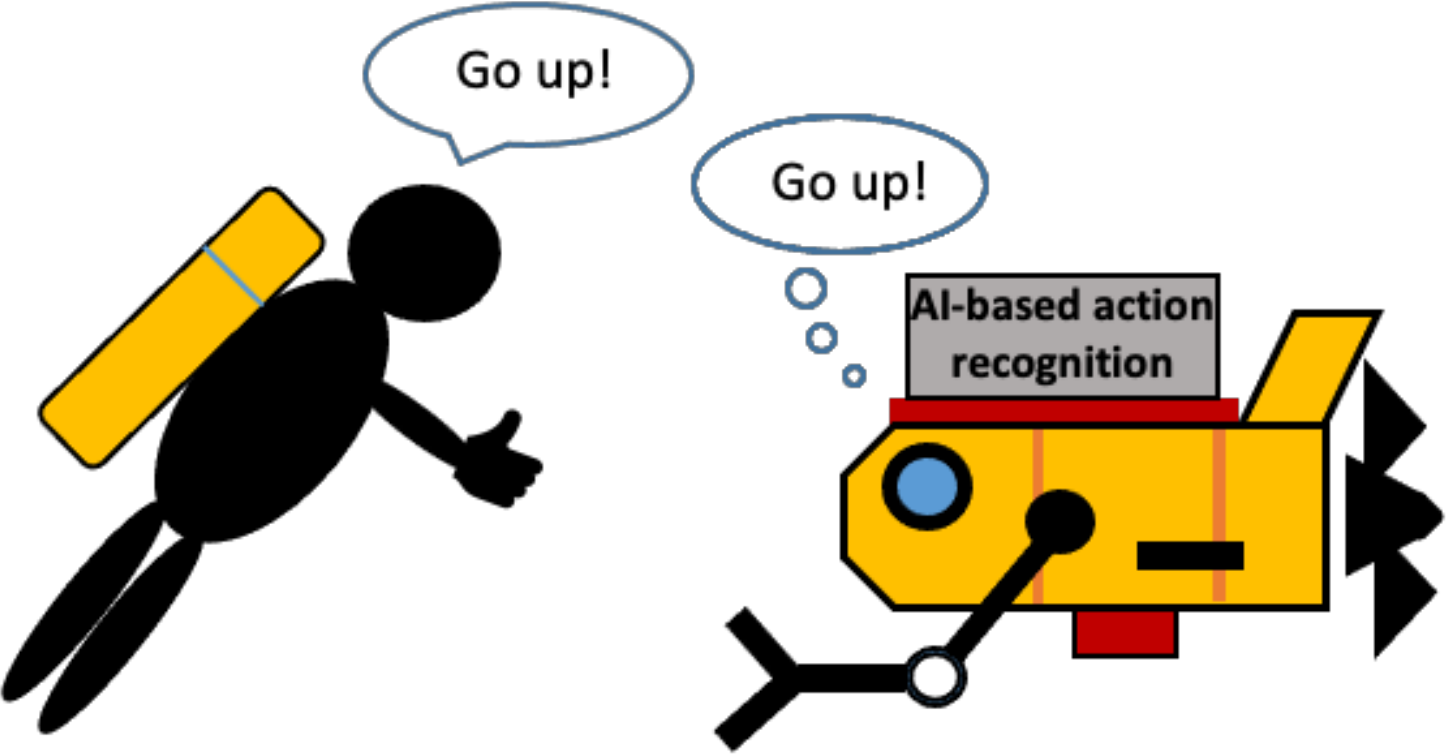} 
\caption{Underwater diver AUV interaction} 
\label{fig:HRI} \vspace{-9pt}
\end{figure}

Since on-demand reprogramming of the AUVs is difficult in dynamic and uncertain underwater environments, they require constant supervision from the diver to perform tasks. Traditional methods of communicating and controlling the AUVs utilize waterproof tablets, keyboards, mice, or joysticks directly connected to the AUVs. These interfaces, however: 1) are expensive to waterproof and deploy in underwater environments, 2) require the diver to be close to the AUV, and 3) are cumbersome and unwieldy to operate. Therefore, due to lack of effective technologies of underwater radio and wireless communication, one way for the diver to send commands to an AUV is by means of hand gestures~\cite{Bandeira2015_gesturebased_controlrobots}. In response, the AUV can utilize AI to recognize the diver's hand gesture and interpret the corresponding command, as seen in Fig.~\ref{fig:HRI}. Sometimes the AUVs are also required to follow the diver by recognizing diver's posture. It is desired that such diver action recognition system implemented on an AUV should be fast to act in time-critical situations, such as changing environments, risky diver locations, and limited oxygen supply.

\begin{figure*}[t] 
\centering
\includegraphics[width=0.98\textwidth]{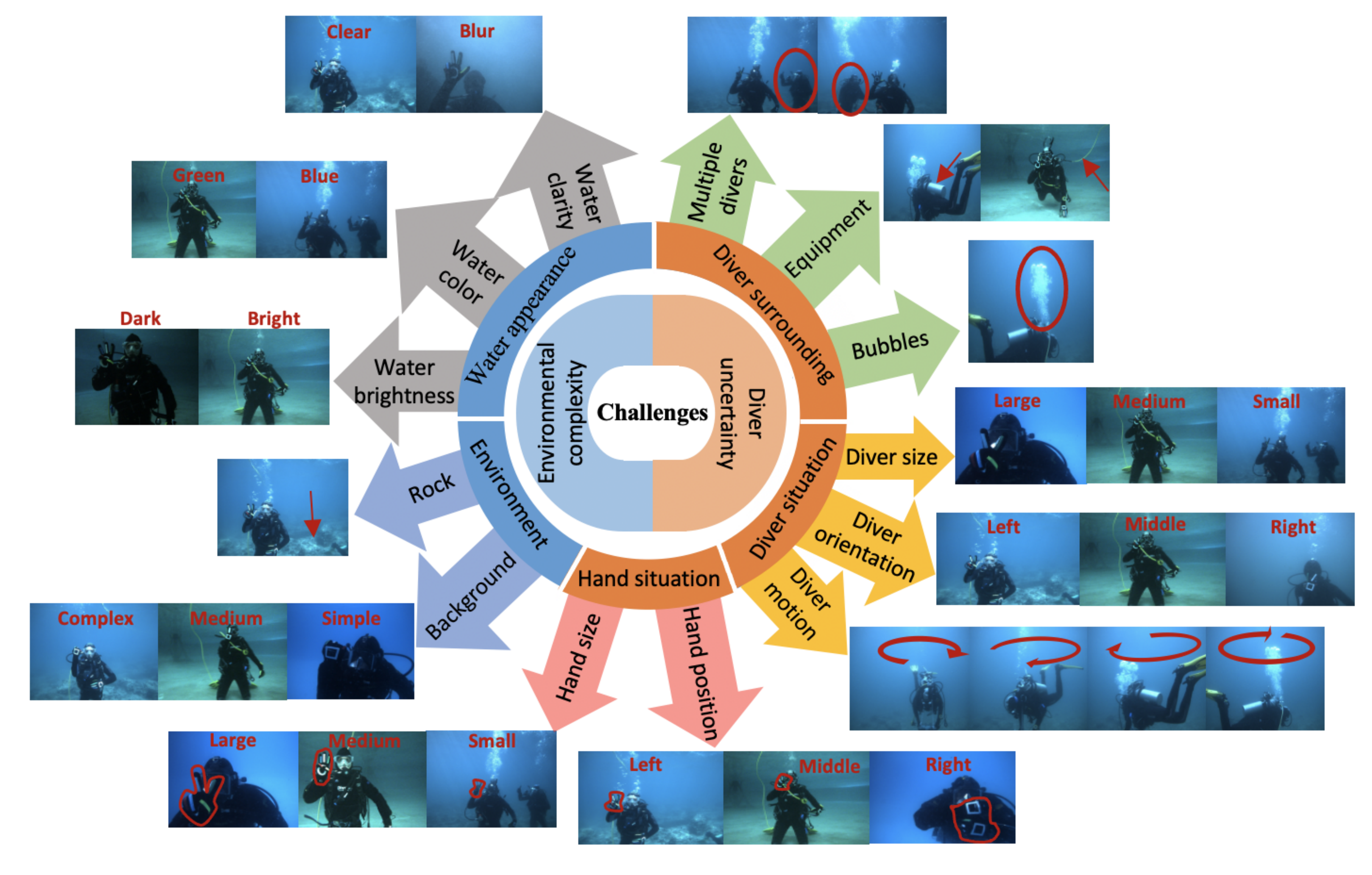} \vspace{-8pt}
\caption{Various challenges of the diver action recognition problem in complex underwater environments} 
\label{fig:challenges} \vspace{-9pt}
\end{figure*}

\vspace{-6pt}
\subsection{Challenges}
There exist several challenges in building a robust, reliable and real-time underwater diver action recognition system, as highlighted in Fig.~\ref{fig:challenges}. These challenges are discussed below.

\vspace{3pt}
\subsubsection{Environmental Uncertainties}
The two main challenges in underwater environments are due to: a) the water appearance and b) the background. Water appearance can adversely affect the image quality due to varying levels of water clarity, color, and brightness.  This results in a large combination of contrasting environments that complicate the diver action recognition problem; thus requiring different image enhancement and filtering techniques to recognize diver actions. While the diver action recognition system might need minimal pre-processing in clear, colorless, and bright water environments, it might require complex nonlinear filtering and feature extraction in murky, brown, and dark water environments. 

The difficulties of an underwater diver action recognition system are further exacerbated by the presence of background complexities. These are of three main categories: terrain and rock formations, marine vegetation and wildlife, and artificial constructs. Thus, it is necessary to filter out these background effects for efficient and reliable diver action recognition. 

Therefore, it is critically important that an underwater diver action recognition system must be robust to operating in diverse and complex underwater environments. 

\vspace{3pt}
\subsubsection{Diver Uncertainties}
There are several uncertainties associated with the diver due to sporadic movements and ambiguous actions. The diver's position and motion is usually uncertain in an underwater environment. First, the size of the diver varies depending on the distance from the AUV. If the diver is too close to the AUV, key parts of his body might not be clear within the camera's field of view. On the other hand, if the diver is too far from the AUV, the camera might not capture enough details to recognize his actions. Second, it is impossible for the diver to remain stationary at all times because of the underwater current. Specifically, the diver's orientation changes with respect to the direction of the current. The diver's position can also be anywhere in the image such as at the bottom left, top middle, or center right. Finally, the diver is moving when performing  a particular pose such as turning horizontally, thus making the problem more difficult. 

Similarly, the position and motion of the diver's hand is crucial for gesture recognition. Similar to the diver size and diver orientation, hand size and hand position can affect the recognition of hand gesture. However, unlike estimating the diver's body pose, the hand is a much smaller target, which requires the recognition model to extract and process particular regions of interest in the image. 

Furthermore, the problem is exacerbated by the presence of bulky equipment (e.g., oxygen tank) carried by the divers. In addition, the exhalation of air bubbles into the underwater environment adds considerable disturbance in the images. Other issue arises, when multiple divers appear in front of the AUV camera system at the same time; thus making it difficult to identify the diver issuing the commands.  

Therefore, it is vital for the underwater diver action recognition system to distinguish between different body poses and hand gestures in the presence of various diver uncertainties while the diver performs complex maneuvers at various positions and orientations relative to the AUV.

\vspace{3pt}
\subsubsection{Sensing Uncertainties}
The underwater camera system is susceptible to noise and biases, which might adversely affect the clarity and color accuracy of images. In addition, various objects in the environment can interfere with the camera's vision.  For example, floating underwater debris near the lens, particles scattered throughout the environment, and reflective underwater surfaces can result in viewing angle obstruction, difficulty focusing the camera lens, and lens flare. These make it difficult to determine if the image distortion is due to the environment, diver motion, or hardware issues. 

Thus, it is important that the underwater diver action recognition system is robust to these sensing uncertainties. 

\vspace{3pt}
\subsubsection{Fusion of Stereo Camera Images}
The training dataset used in this paper consists of images that are captured using a stereo camera system, which contains one camera with two separate lenses abreast in order to implicitly preserve the distance information. This system uses the left-lens and the right-lens to capture two images of a diver action at the same time from  slightly different angles. If only the left or the right lens image is used to classify the diver action, then it might exclude the crucial distance information and thus it can degrade the classifier performance. On the other hand, treating left and right-lens images as individual uncorrelated images in the same training dataset will cause the model to over-fit since these images correspond to the same diver action. Thus, the left and right lens images should be fused together at the feature level prior to classification. Fusion, however, is difficult because of stereo correlation and subtle spatial differences. 

Therefore, the underwater diver action recognition system should be able to fuse the left and right lens information cohesively to boost the classification accuracy. 

\vspace{3pt}
\subsubsection{Computational Efficiency and Reliability}
While performing tasks in hazardous underwater environments, there may be several time-critical situations where commands must be carried out by the AUV urgently to ensure the safety of the diver and/or the AUV, and to accommodate the changing environment. However, it might be difficult to interpret commands fast because of the high-resolution RGB stereo imagery captured by the camera system and limited computational resources on the AUV. Furthermore, many frames may need to be processed in rapid succession to verify a sequence of commands to be executed.  
Additionally, it is required that the diver actions are recognized with high accuracy to ensure that the AUV operates reliably in time-critical situations. 

Therefore, the on-board diver action recognition system needs to be simple for ease of implementation, accurate for robust and reliable performance, and computationally efficient for classification of diver actions in real-time.

\vspace{-9pt}
\subsection{Related work}
Remotely operated vehicles (ROVs) are common underwater robots which are directly controlled by a human operator from the surface via cable connections. These vehicles, however, are limited in their deployment and autonomy due to expensive equipment,  short range, unwieldy cables, and  complex operation~\cite{Capocci2017_ROV_review}. On the other hand, autonomous underwater vehicles (AUVs) are cable-free, equipped with advanced sensing and control technologies~\cite{Nicholson2008}, provide longer range autonomy and are becoming increasingly more intelligent and cost effective. Thus, they have become useful to serve as advanced assistants for a variety of underwater tasks. However, despite the advances in AUV technologies, several underwater tasks require efficient diver-AUV collaboration. The key requirement for diver-AUV collaboration is the capability to dynamically reprogram the AUV's task parameters. In underwater environments, radio and wireless communications between the diver and AUVs\cite{Dudek2008_Sensorbased_AUV} are not efficient. Traditionally, divers used tactile devices to send commands using waterproof tablets, keyboards, mice and joysticks; however,  these interfaces are expensive and inconvenient~\cite{Bandeira2015_gesturebased_controlrobots}. Therefore, the underwater diver action recognition problem has become an increasingly important area of research to enable efficient underwater diver-AUV interaction. 

There have been some initial attempts to address the diver action recognition problem, which requires: 1) the creation of a database of diver actions in diverse and realistic underwater environments, 2) the use of cost-effective but information-rich underwater sensors to capture diver motions, and 3) a computationally efficient and reliable diver action recognition model.  Initial studies focused on the classification of diver actions were based on images captured in ideal swimming pool environments \cite{Islam2018_Dynamicreconfiguration_humanrobot_interactive} with controllable lighting and background conditions. These were later extended to include diver actions in more realistic underwater ocean environments \cite{Islam2018_Understandinghumanmotion_AUV}. In both cases, images were captured using a single monocular RGB camera. While this sensing solution is cost-effective, this system has several limitations including blind spots from a single point-of-view, low image resolution, and limited depth information which can make the system unreliable.

The diver pose recognition problem is also crucial for diver tracking application~\cite{Miskovic2015}. Traditionally, acoustic sensors are used to locate the diver's position relative to the water surface for autonomous surface vehicles; however, such sensors cannot identify specific diver poses (e.g., orientations and motion types). Similarly, information-rich sonar, ultra-short baseline acoustic localization, and stereo-cameras have also been used for AUV diver tracking. These sensors ensure the AUV closely follows the diver underwater; however,  diver action recognition models were not developed to identify specific diver poses present in the captured images\cite{Nad2020}.  At the same time, research has also focused on detecting the number of divers present in the view of the AUV camera in various underwater environments \cite{Islam2018_Understandinghumanmotion_AUV,Islam2018_motion}. Researchers have also addressed the diver arm motion recognition problem using wireless acoustic networks in a simulated underwater environment \cite{hu2020}; however, there is no real-world data for validation. Furthermore, these networks might not always be available for implementation in the ocean environments. 

Common underwater recognition models utilize time-frequency feature extractors  with k-nearest neighbor and SVM classifiers to identify mammals and inanimate objects on the ocean floor \cite{Huynh1998,Murugan2017,Kumar2015}. Researchers performed gesture recognition by fusing convex hull method and SVM classifier~\cite{Gustin16}; however, very few gestures and no poses were considered. As such, the method might not scale well to larger and more complex set of diver actions. On the other hand, deep transfer learning methods have been applied to underwater fish species classification \cite{Siddiqui2018}, human motion recognition in terrestrial environments \cite{wang2018} and diver gesture recognition\cite{JYang2019}. While these methods provide good average performance, high individual class performance is not guaranteed. 

However, for the diver action recognition problem it is of critical importance that each diver action is recognized with high accuracy for safe and reliable operation in underwater environments. To the best of our knowledge, the diver action recognition problem is still an active area of research.

\vspace{-6pt}
\subsection{Contributions}
To address the challenges and limitations discussed above, this paper develops a robust, reliable and computationally efficient Diver Action Recognition System, called DARE, to identify the diver hand gestures and full-body poses in real-time; thus, enabling the AUVs to reliably interpret diver commands. DARE has been trained based on a recently created rich database of diver hand gestures and poses, called Cognitive Autonomous Diving Buddy (CADDY) \cite{Chavez2019_CADDY} dataset, which contains data generated in real uncertain ocean environments.  

DARE is built upon deep transfer learning architecture for robust classification of a multitude of diver actions under various underwater uncertainties and complex scenarios. 
First, DARE enables fusion of information available from the left and right images of the stereo camera system on AUVs. For this purpose, the deep transfer learning method is extended to a multi-channel framework that: 1) extracts the relevant features from each stereo image individually using the convolution layers, and 2) optimally fuses these features together using fully connected neural networks.

Second, DARE provides high classification performance of each individual class. Typically, a single monolithic neural network classifier is trained after the convolution layers; however, for problems with large number of classes, such networks cannot guarantee a high minimum performance for each individual class. Therefore, instead of considering all diver actions simultaneously, DARE trained a decision tree of fully-connected neural network classifiers, where each network is tailored to discriminate between groups of diver actions. This ensures high individual class  recognition performance at the bottom of the tree, thus yielding superior reliability.

The main contributions of this paper are as follows:

\begin{itemize}
  
  \item Development of DARE, to facilitate computationally efficient, robust, reliable and highly accurate diver action recognition, under various environmental, diver, and sensing uncertainties. It consists of:
  \begin{itemize}
      \item  a multi-channel deep transfer learning architecture for fusing stereo pairs of camera images and
      \item a hierarchical tree structured classification scheme to yield high individual class recognition performance. 
  \end{itemize} 
    \item Training and testing of DARE using the CADDY dataset consisting of a large number of diver action images in real-life challenging underwater environments.
    \item Comparative evaluation with baseline architectures.
\end{itemize}

\begin{table*}[t!]
\caption {CADDY dataset diver hand gestures}\label{tbl:gestures caddy} \vspace{-6pt}
 \begin{center}
\footnotesize
\bfseries
    \begin{tabular}{|l|l|l||l|l|l||l|l|l|}
     \toprule
    \tabincell{l}{Diver Image} & \tabincell{l}{Gesture} & \tabincell{l}{Code} & \tabincell{l}{Diver Image} & \tabincell{l}{Gesture} & \tabincell{l}{Code} & \tabincell{l}{Diver Image} & \tabincell{l}{Gesture} & \tabincell{l}{Code} \\
   \hlx{vhvvv} 
     \tabincell{l}{\includegraphics[width=0.105\textwidth]{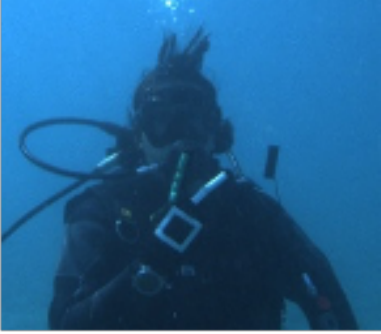}}
      & 
   \tabincell{l}{\includegraphics[width=0.05\textwidth]{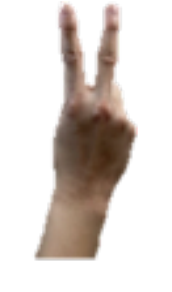}}
      &
 \tabincell{l}{Start}
      &
    \tabincell{l}{\includegraphics[width=0.105\textwidth]{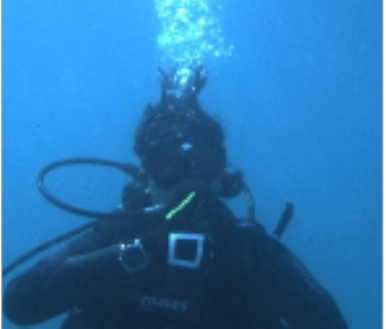}} 
      & 
     \tabincell{l}{\includegraphics[width=0.06\textwidth]{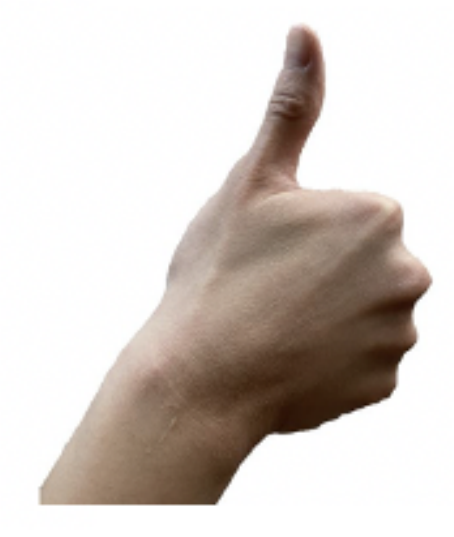}} 
      & 
      \tabincell{l}{Up} 
      &
     \tabincell{l}{\includegraphics[width=0.105\textwidth]{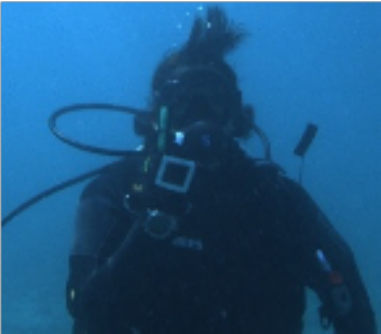}}
      & 
     \tabincell{l}{\includegraphics[width=0.04\textwidth]{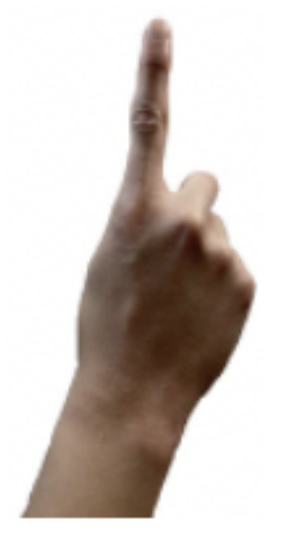}}
      & 
      \tabincell{l}{End}
     \\
     \hlx{vv}
     \tabincell{l}{\includegraphics[width=0.105\textwidth]{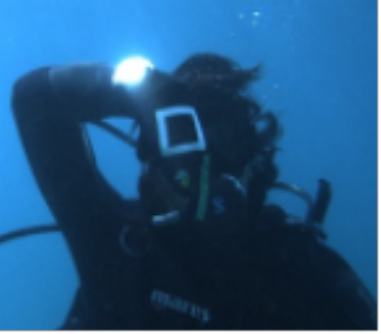}}
      & 
     \tabincell{l}{\includegraphics[width=0.05\textwidth]{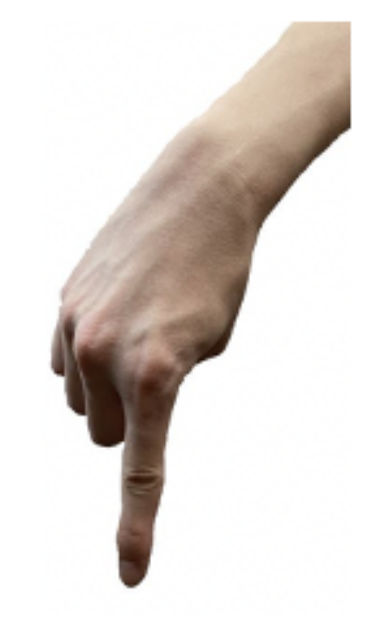}}
      & 
      \tabincell{l}{Here}
     &
     \tabincell{l}{\includegraphics[width=0.105\textwidth]{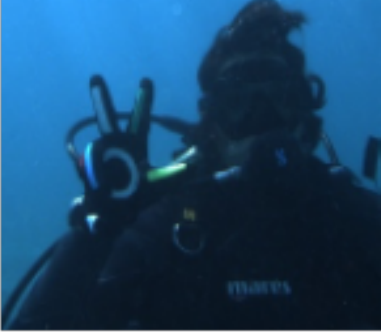}}
      & 
     \tabincell{l}{\includegraphics[width=0.05\textwidth]{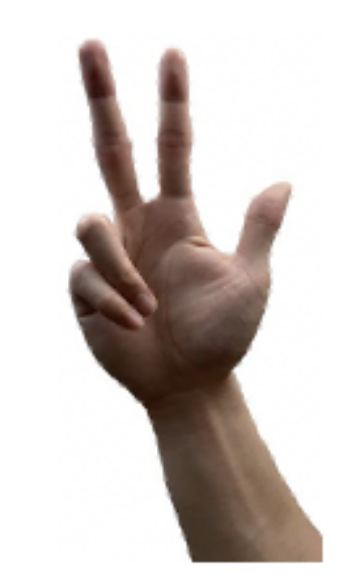}}
      & 
     \tabincell{l}{Take a\\ photo}
      &
     \tabincell{l}{\includegraphics[width=0.105\textwidth]{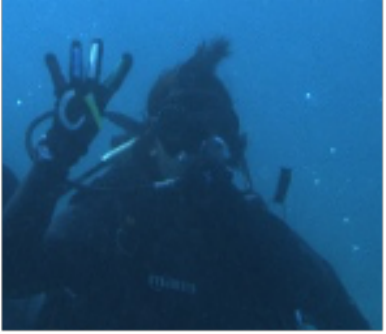}}
      & 
     \tabincell{l}{\includegraphics[width=0.05\textwidth]{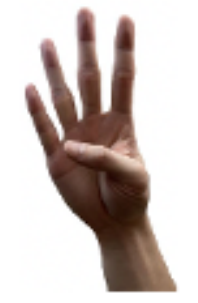}}
      & 
      \tabincell{l}{Four}
      \\
      \hlx{vv}
      \tabincell{l}{\includegraphics[width=0.105\textwidth]{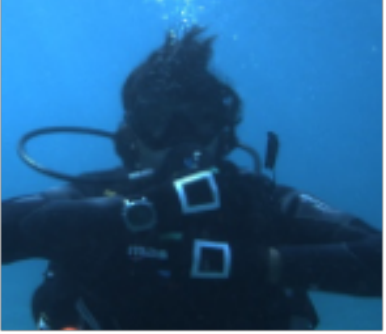}}
      & 
      \tabincell{l}{\includegraphics[width=0.08\textwidth]{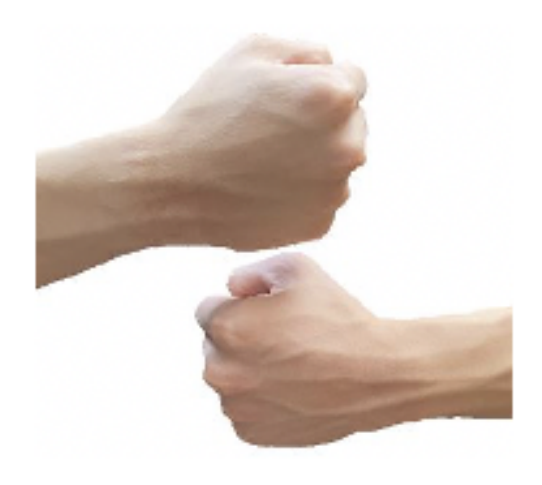}}
      & 
      \tabincell{l}{Carry}
      &
      \tabincell{l}{\includegraphics[width=0.105\textwidth]{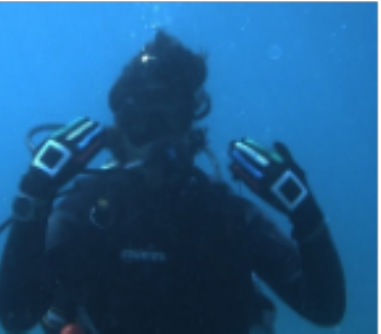}}
      & 
      \tabincell{l}{\includegraphics[width=0.08\textwidth]{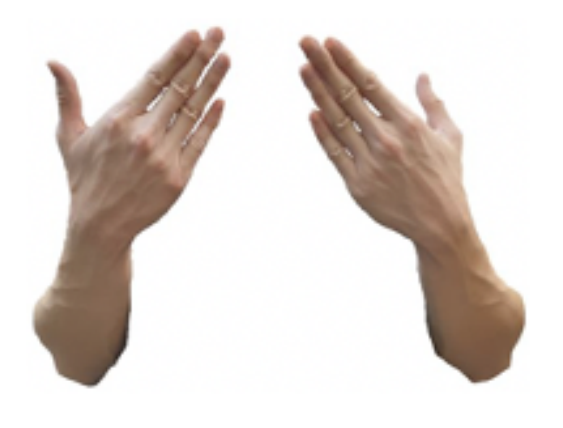}}
      & 
       \tabincell{l}{Tessel-\\lation}
      &
     \tabincell{l}{\includegraphics[width=0.105\textwidth]{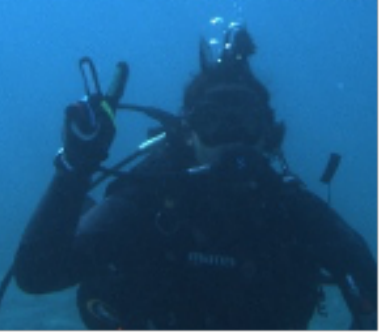}}
      & 
     \tabincell{l}{\includegraphics[width=0.05\textwidth]{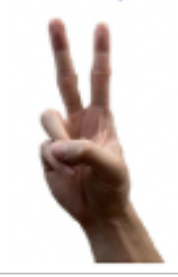}}
      & 
     \tabincell{l}{Two}
     \\
     \hlx{vv}
     \tabincell{l}{\includegraphics[width=0.105\textwidth]{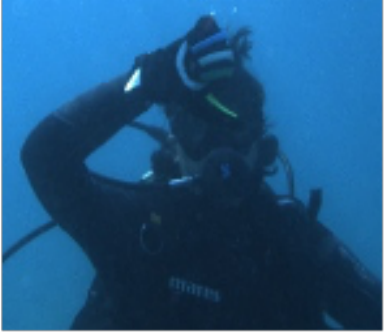}}
      & 
     \tabincell{l}{\includegraphics[width=0.06\textwidth]{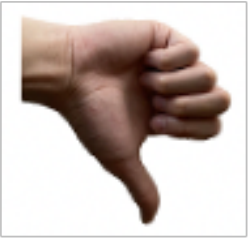}}
      & 
     \tabincell{l}{Down}
      &
     \tabincell{l}{\includegraphics[width=0.105\textwidth]{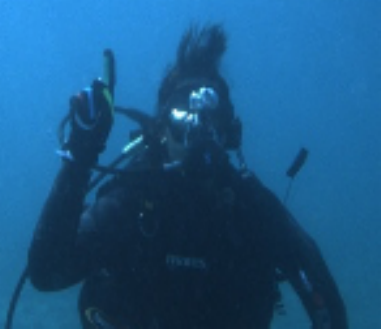}}
      & 
     \tabincell{l}{\includegraphics[width=0.05\textwidth]{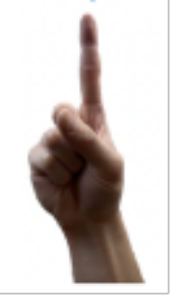}}
      & 
      \tabincell{l}{One}
      &
     \tabincell{l}{\includegraphics[width=0.105\textwidth]{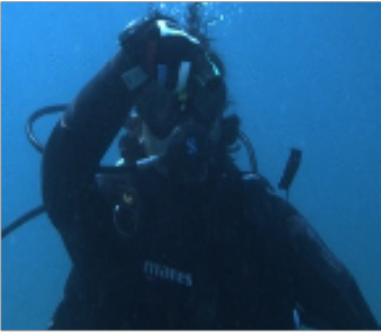}}
      & 
     \tabincell{l}{\includegraphics[width=0.06\textwidth]{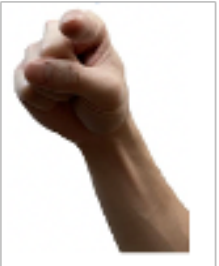}}
      & 
    \tabincell{l}{Back-\\ward}
      \\
      \hlx{vv}
     \tabincell{l}{\includegraphics[width=0.105\textwidth]{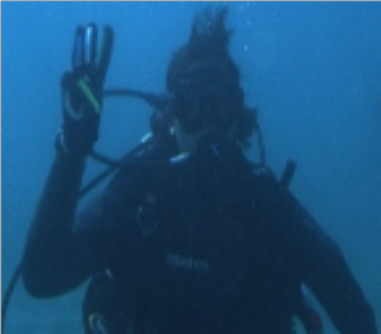}}
      & 
     \tabincell{l}{\includegraphics[width=0.05\textwidth]{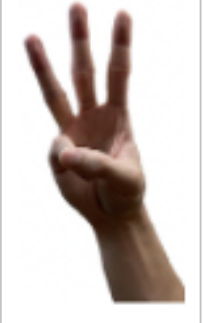}}
      & 
      \tabincell{l}{Three}
      &
     \tabincell{l}{\includegraphics[width=0.105\textwidth]{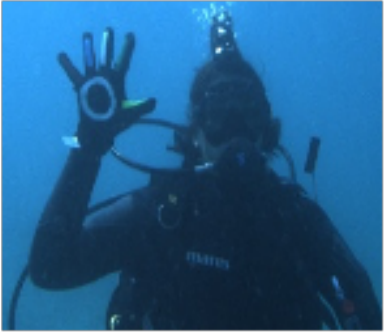}}
      & 
     \tabincell{l}{\includegraphics[width=0.06\textwidth]{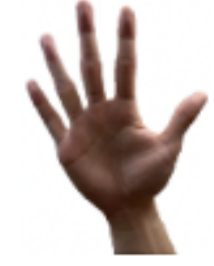}}
      & 
      \tabincell{l}{Five}
      &
     \tabincell{l}{\includegraphics[width=0.105\textwidth]{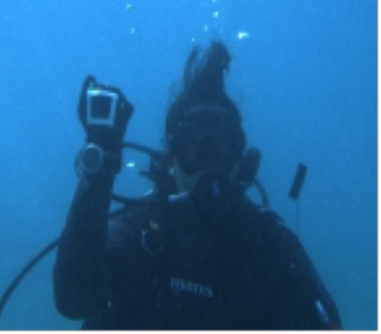}}
      & 
     \tabincell{l}{\includegraphics[width=0.04\textwidth]{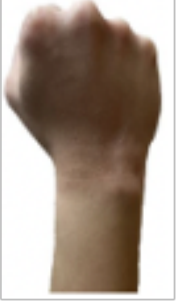}}
      & 
      \tabincell{l}{Number\\delimiter}
      \\
      \hlx{vv}
      \tabincell{l}{\includegraphics[width=0.105\textwidth]{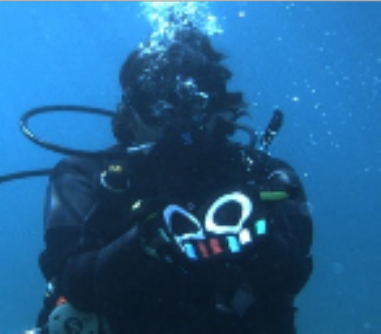}}
      & 
     \tabincell{l}{\includegraphics[width=0.07\textwidth]{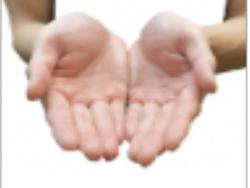}}
      & 
     \tabincell{l}{Boat}
      &
      &
      &
      &
      &
      &
     \\
     \bottomrule
      \end{tabular}
    \end{center} \vspace{-6pt}
      \end{table*}

\vspace{-3pt}
\subsection{Organization}
The rest of the paper is organized as follows: Section~\ref{sec:dataset} describes how CADDY dataset was created and organized for both gestures and poses. Section~\ref{sec:method} provides the details of the deep transfer learning tree structure approach for classifying the diver gestures and poses. Section~\ref{sec:result} presents the data analysis setup, compares the performance with other existing techniques, and discusses the results. Finally, Section~\ref{sec:conclusion} presents the conclusions and plan for future work.

\section{CADDY Dataset Overview}
\label{sec:dataset}
The objective of an underwater diver action recognition system is to enable remote supervision of AUVs to remove the need of cumbersome and expensive waterproof joysticks and keyboards. Such a system should have low complexity for real-time execution once deployed on an AUV. Furthermore, it should be robust to various underwater uncertainties (e.g., environment, diver, and sensor uncertainties) and deliver highly reliable classification performance. 

Therefore, to train such a system this paper utilizes the Cognitive Autonomous Diving Buddy (CADDY)\cite{Chavez2019_CADDY} dataset, which consists of a rich variety of diver action images collected in real uncertain underwater environments. Specifically, images in the CADDY dataset were captured using the BumbleBee XB3 stereo RGB camera system with a resolution of $640 \times 480$ pixels. The stereo camera is similar to human eyes which contains one camera and two lenses side by side; it takes synchronized pictures which together preserve the information like spatial distance, object/diver details and the broader view of the background environment\cite{Shortis2016}. These images were collected in  various indoor swimming pools and in the open seas under complex and diverse environmental conditions. Specifically, the images were collected in three different locations: 1) the open seas of Biograd na Moru, Croatia, 2) an indoor pool in the Brodarski Institute, Croatia, and 3) an outdoor pool in Genova, Italy. Images collected from these  distinct locations have diverse conditions such as water color, water clarity, underwater lighting,  terrain, diver equipment, diver movement, size of the gestures and their position and orientation. All these factors make the diver gesture and pose recognition problem very challenging.

The CADDY dataset consists of: 1) diver hand gestures to demonstrate specific commands for the AUV under realistic underwater operating conditions and 2) diver poses for the  AUV to track and follow the diver.

\vspace{-6pt}
\subsection{Diver Gestures}
Hand gestures in the CADDY dataset  use the CADDIAN gesture-based language\cite{Chiarella2018}. The diver signals a specific hand gesture or performs a sequence of gestures  to command the AUV in real-world underwater missions. In order to  aid the classifier in distinguishing between different hand gestures, the divers wear gloves with special features that have a different color strip on each finger, a white square on the back of the hand, and a white circle on the front of the hand. Without these features on the gloves, it is not only difficult to distinguish between different hand gestures but also difficult to differentiate between the diver's body and his hands in dim surroundings even with human eyes.

Table~\ref{tbl:gestures caddy} shows a total of 16 different gestures and the associated commands, such as start, up, down, and take a photo. 
Overall, the CADDY dataset contains $9239$ stereo pairs, i.e., $9239\times2=18,478$ images of the diver gestures.

\vspace{-6pt}
\subsection{Diver Poses}
Pose images were extracted from video sequences showing the diver's movement. The purpose of classifying diver's pose is to track diver's movement and position such that the diver is always facing the camera on the AUV. Table \ref{tbl:posecaddy} shows $3$ different poses: i) turning horizontally with chest pointing downward to the floor, ii) turning clockwise or counterclockwise vertically, and iii) free swim. Overall, the dataset contains $12,708\times2=25,416$ images of the diver poses.

In this paper, DARE is trained using both gesture and pose datasets, thus it is capable of recognizing $20$ different classes consisting of $16$ gestures, $3$ poses, and $1$ true negative (or null) class. The null class contains $7190$ stereo pair which include miscellaneous situations like diver idle, diver missing, gestures missing, improper gestures or the transition between two gestures. Thus, there are a total of $(9239+12,708+7190)\times2=58,274$ images which are used for training and validation of DARE. 

\begin{table}[t]
\caption{CADDY dataset diver postures}
      \label{tbl:posecaddy} \vspace{-6pt}
     \begin{center}
     \footnotesize
     \bfseries
     \begin{tabular}{c|c|c|c}
     \toprule
      Command & Turning & Turning  & Free  \\ 
      & horizontally & vertically & swim \\ 
      \hlx{vhvv}
     \tabincell{c}{Diver\\Posture}
     &
     \tabincell{c}{\includegraphics[width=0.11\textwidth]{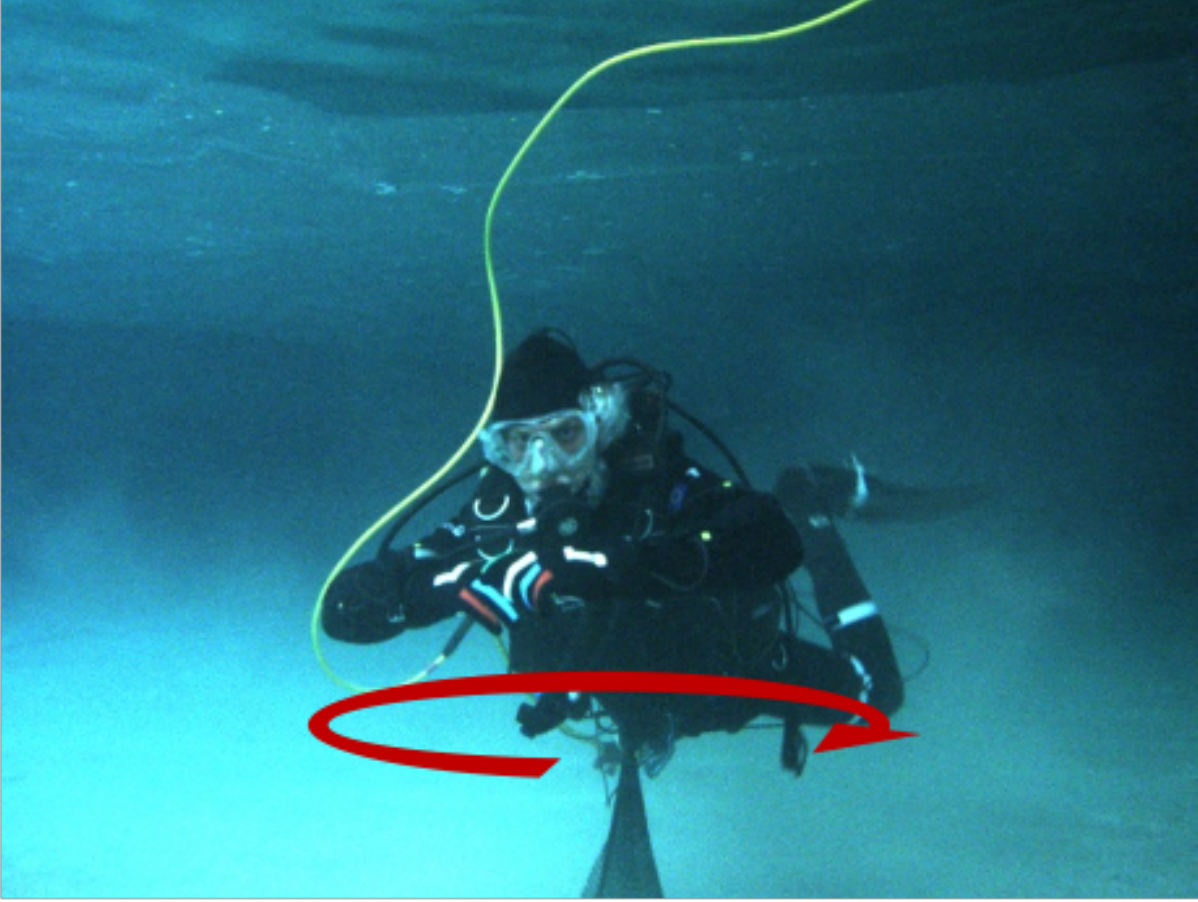}}
     & 
     \tabincell{c}{\includegraphics[width=0.11\textwidth]{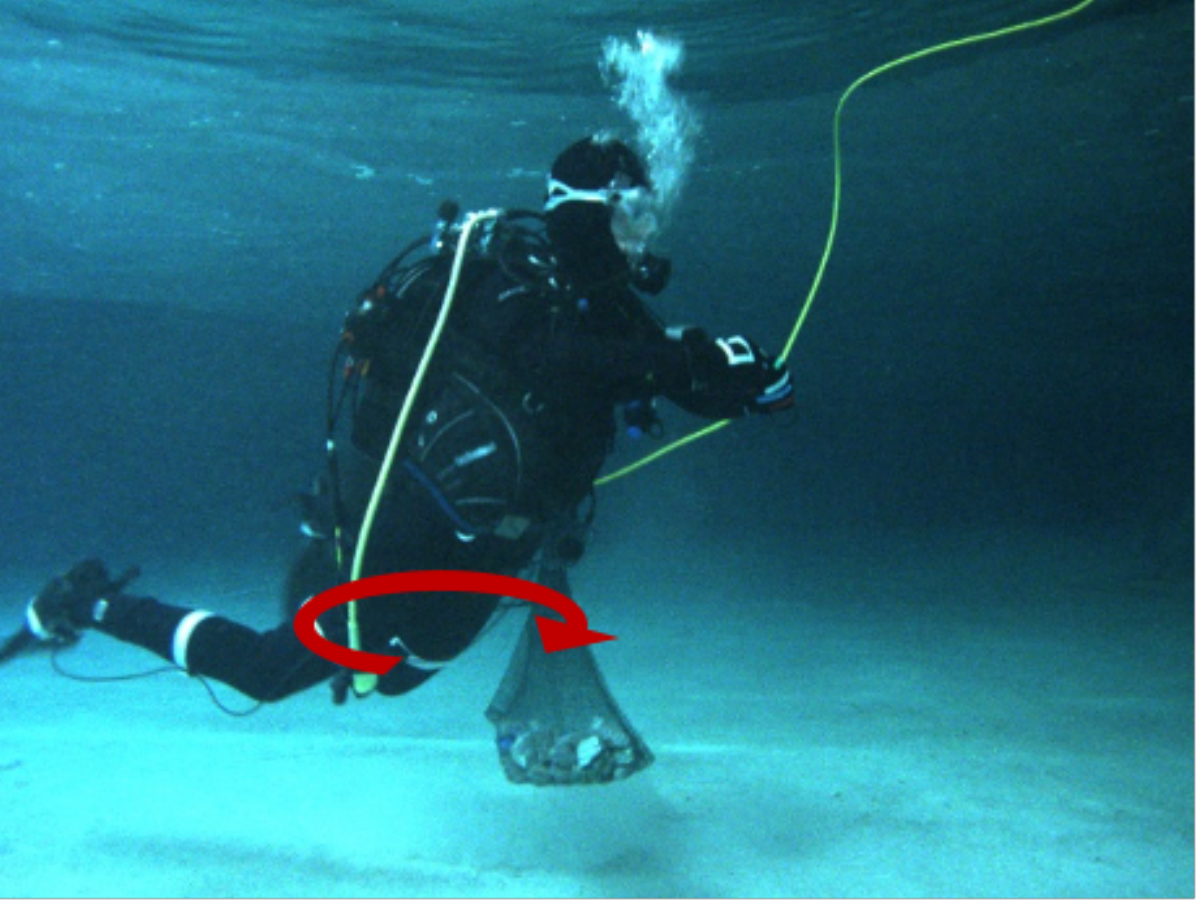}}
     & 
     \tabincell{c}{\includegraphics[width=0.11\textwidth]{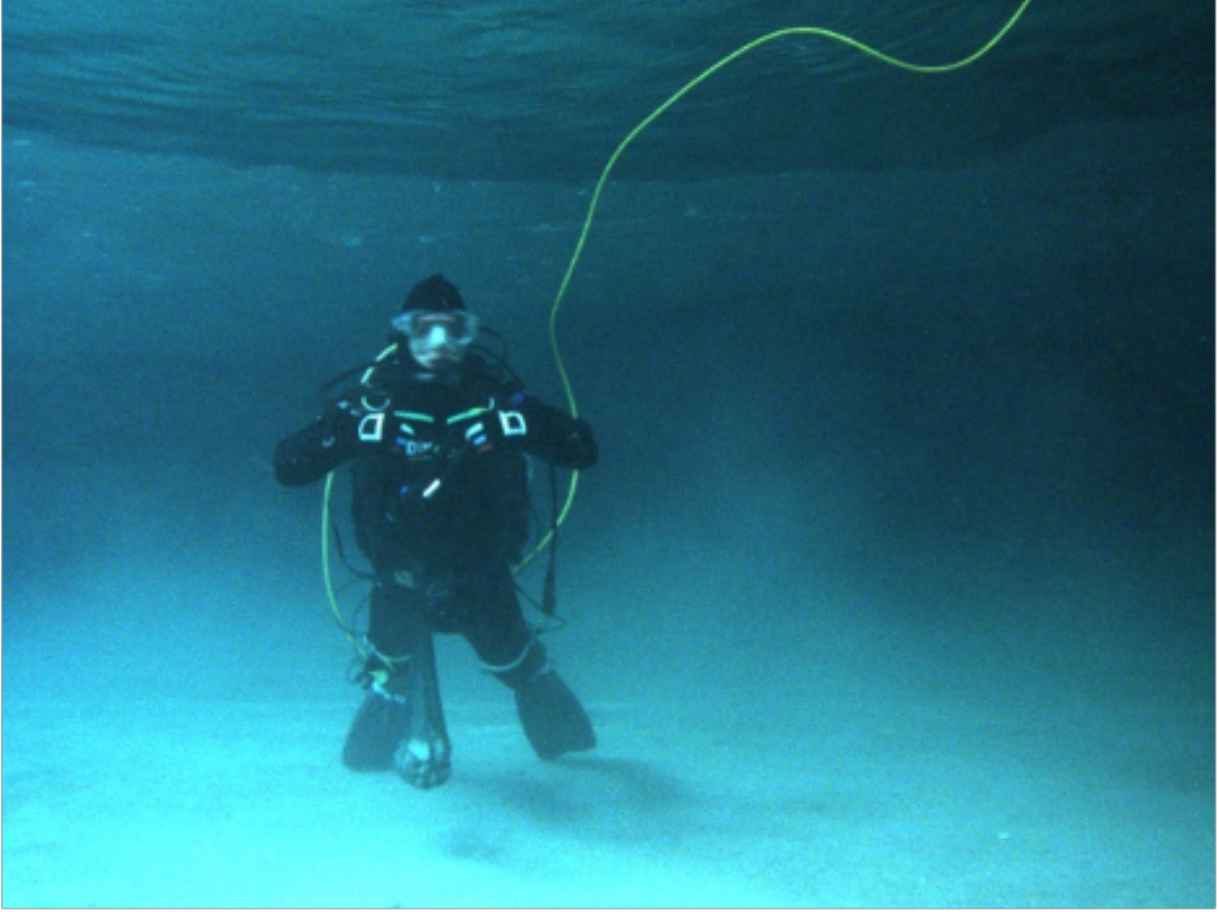}}
     \\ \bottomrule
      \end{tabular}
      \end{center}\vspace{-6pt}
\end{table}

\section{DARE Architecture}
\label{sec:method}
\begin{figure*}[t!] 
\includegraphics[width=1.0\textwidth]{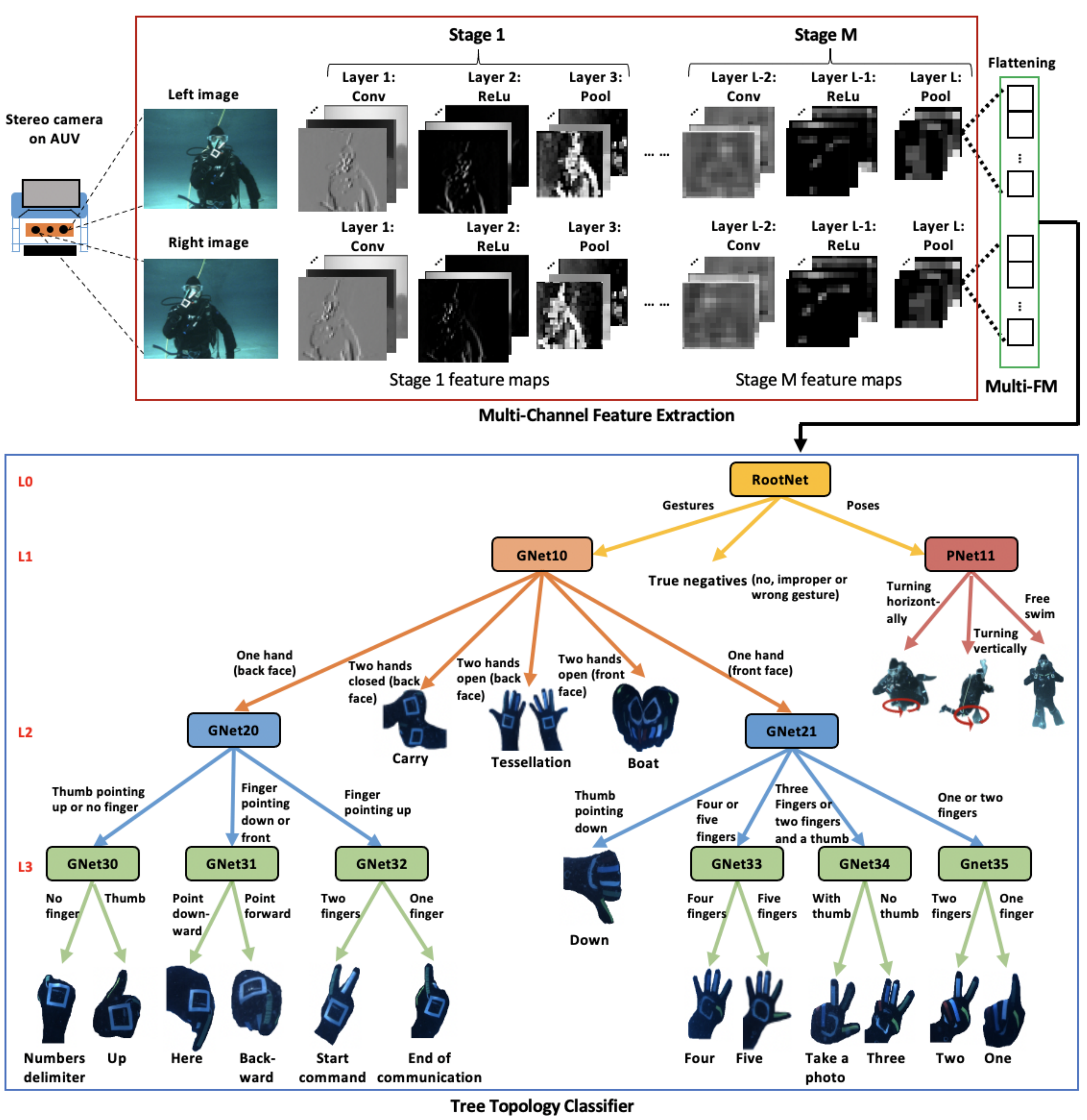}
\caption{DARE architecture}
\label{fig:training progress}\vspace{-12pt}
\end{figure*}

DARE is constructed to enable robust, efficient and reliable underwater communication between the diver and AUV considering various environmental, diver, and sensing uncertainties. Specifically, it is critical that any diver action is accurately recognized in real-time in order to enable safe and rapid completion of tasks in hazardous underwater environments. The DARE architecture is shown in Figure~\ref{fig:training progress}. During AUV missions, a live video of the divers is recorded using an on-board stereo camera. Each frame captured in the video is composed of two images, corresponding to the left and right lens of the stereo camera. Deep multi-channel transfer learning architectures are utilized to extract the features from each image individually. These features from each image are then "flattened" into a single one-dimensional feature vector, which is then systematically fused and classified to identify the diver action using a tree structured classifier consisting of fully-connected neural networks. The tree-structure helps boost the individual class recognition performance.

\begin{figure*}[t!]  
\begin{centering}
\includegraphics[width=0.95\textwidth]{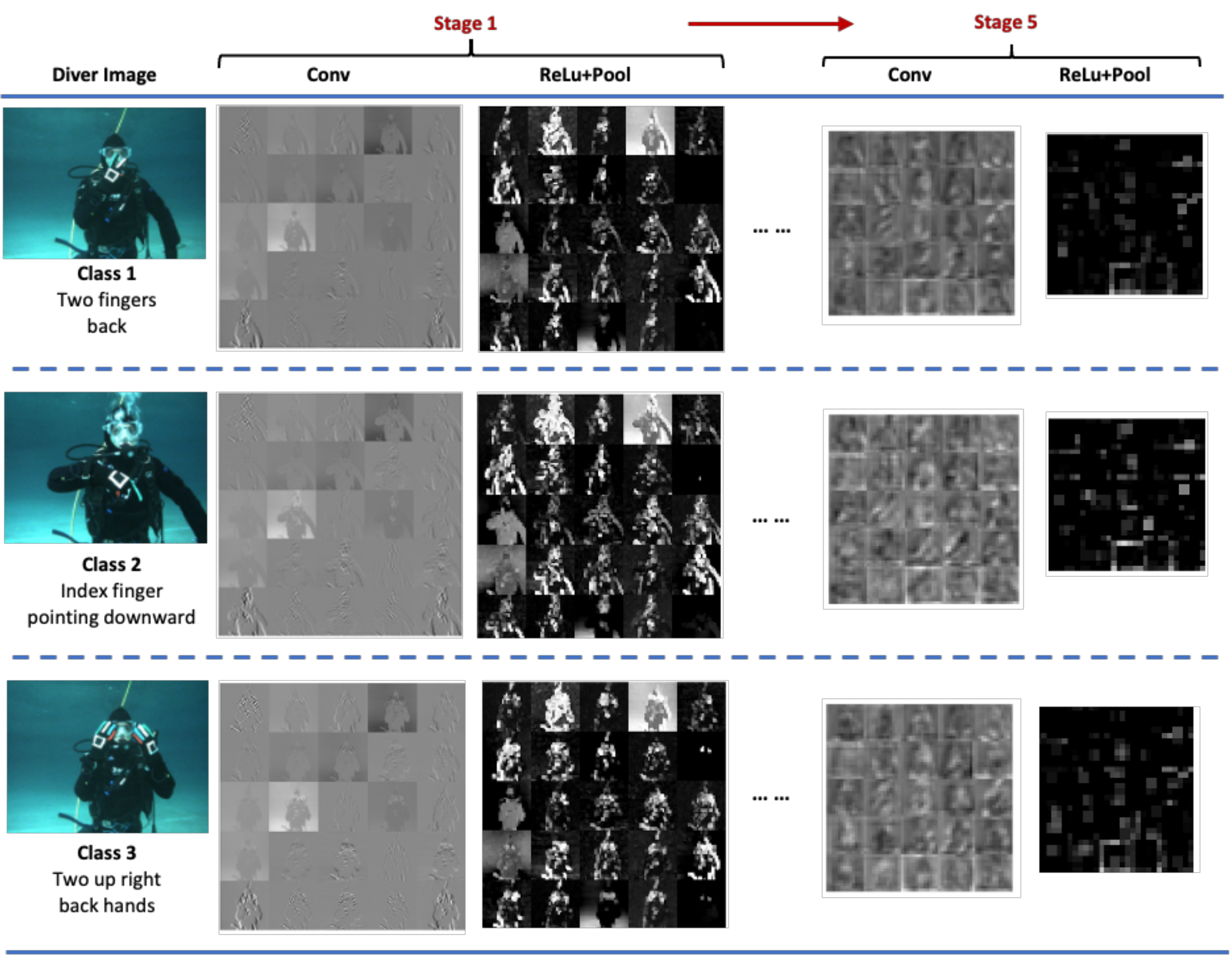}
\caption{Feature visualization with selected features.}
\label{fig:feature}
\end{centering}
\vspace{-6pt}
\end{figure*}

\subsection{Multi-Channel Feature Extraction}
DARE is built upon the deep transfer learning approach utilizing CNNs in a bi-channel configuration (i.e., one for each image of the stereo pair of the AUV camera). 

Multi-channel CNN approaches have been applied in other domains that have multiple high-dimensional input data streams for robust and accurate classification \cite{Wu19,Jiang19}. In multi-channel frameworks, data from each input stream is processed independently before fusion and classification. In the context of stereo camera system in the CADDY dataset, we employ a bi-channel approach to enable feature extraction from two different perspectives, which: i) implicitly capture the distance of the diver from the AUV, and ii) allow for better filtering of noise, biases, and other uncertainties by comparing correlated details between stereo images. Thus, a bi-channel CNN approach enables DARE to take full advantage of stereo images and provide robust classification.

The training of deep CNNs requires a large dataset (on the order of tens of thousands of images per class) in order to adequately train the weights of the network without overfitting.  However, the collection of such large amounts of data is often cost-prohibitive and requires expensive computational resources to process. In order to overcome these limitations, we employ the use of pre-trained deep transfer learning models\cite{Pan2009}. These models were trained on large-scale and diverse benchmark datasets, and their structure can be transferred to solve a different, smaller, but also relevant problem. In a transfer learning approach, the weights of the pre-trained convolutional layers are directly utilized to extract the features of the input images, and only the fully-connected neural network layers are retrained. This approach therefore only requires minimal data for re-training and thus shorter computation times, while still yielding high classification accuracy.

In this paper, DARE is constructed using three basic pre-trained transfer learning models: ResNet18 \cite{He2016_Resnet}, AlexNet \cite{Krizhevsky2012_Alexnet}, and VggNet16 \cite{Simonyan2014_Vggnet}. The architectures of these models are summarized in Table~\ref{tab1}. In general, these models are composed of an input layer, a series of convolutional stages that extract abstract and relevant features from the images, and finally a fully connected neural network that fuses the features and makes a classification decision. The input images from the camera are down-sampled to match the input size requirements of the considered transfer learning model.

Both channels are identical to each other and consist of the same stages as the underlying transfer learning model. These stages typically consist of a convolutional layer with a specified filter size that extracts the features, an activation layer that augments the features via nonlinear transformations to help increase separation, and finally a pooling layer that downsamples the features to ensure they are compact to maintain generality while reducing computation times. An example of the feature extraction process from the AlexNet is shown in Figure~\ref{fig:feature}. The final features of each channel are then flattened and concatenated before being sent to the tree topology classifier for robust information fusion and decision making. For the sake of completeness, a brief overview of the most common CNN layers in each stage of a deep transfer learning model~\cite{DLTutorial} is presented below.

\begin{table}[t!]
\normalsize	
\caption{Specifics of transfer learning models}\vspace{-6pt}
\begin{center}
\begin{tabular}{c|c|c|c}
\hline
&\textbf{ResNet}&\textbf{AlexNet} & \textbf{VggNet}\\
\hline
\tabincell{c}{Network depth} &  \tabincell{c}{18} & \tabincell{c}{8} & \tabincell{c}{16} \\
\hlx{v}
\tabincell{c}{Input layer\\ size} & \tabincell{c}{224$\times$224} & \tabincell{c}{227$\times$227} &\tabincell{c}{224$\times$224} \\
\hlx{v}
\tabincell{c}{Convolution\\layers} & \tabincell{c}{17} & \tabincell{c}{5} & \tabincell{c}{13}\\
\hlx{v}
\tabincell{c}{Fully connected\\ layers} & \tabincell{c}{1} & \tabincell{c}{3} & \tabincell{c}{3}\\
\hlx{v}
\tabincell{c}{$\#$ of parameters\\(million)}& \tabincell{c}{11.7} & \tabincell{c}{61.1} & \tabincell{c}{138.4}\\
\hline
\end{tabular}
\label{tab1}
\end{center}\vspace{-18pt}
\end{table}

\subsubsection{Convolutional Layer}
Each convolution layer outputs a feature map generated from a set of filters. Without loss of generality, suppose layer $l$ is a convolution layer. Then, the convolution operations in this layer are as follows. The input to this layer is a 3D matrix $\mathbf{Z}_{l-1}$ of feature maps generated from the previous layer $l-1$ with dimensions $N_{l-1}\times N_{l-1}\times C_{l-1}$, where $N_{l-1}$ is the length and width, and $C_{l-1}$ is the depth (i.e., the total number) of feature maps. Define padding parameter $p_l\in \mathbb{Z}^{+}\cup\{0\}$ as the number of zero-element rows or columns to add to each outer edge of each input feature map.  This padded set of feature maps is denoted as $\mathbf{Z}^{p_l}_{l-1}$ with dimensions $(N_{l-1}+2p_l)\times(N_{l-1}+2p_l)\times C_{l-1}$. The convolution operations are performed on $\mathbf{Z}^{p_l}_{l-1}$. Suppose there are $C_l$ filters in layer $l$ and let $k\in\{0,\dots,C_l-1\}$ be the filter index. Define filter $k$ as a 3D matrix of weights $\mathbf{W}_l^{k}$ with dimensions $V_l\times V_l \times C_{l-1}$, where $V_{l}\leq N_{l-1}+2p_l$. Each filter has an associated scalar bias term $b_l^{k}$. Finally, define stride parameter $s_l\in\mathbb{Z}^+$, which is the step size of the convolution filter both horizontally and vertically across the maps. Then, the 3D output feature map of layer $l$ is computed as: 
\begin{equation}
    \label{eq:convolution}
    \begin{split}
    \mathbf{Z}_l[i][j][k] = b_l^k+\sum_{x=0}^{V_l-1}\sum_{y=0}^{V_l-1}\sum_{c=0}^{C_{l-1}-1}  & \mathbf{Z}^{p_l}_{l-1}[s_l i + x][s_l j + y][c] \\
    & \times \mathbf{W}^k_l[x][y][c] \\
    \end{split}
\end{equation}
where $i,j\in\{0,\dots N_l-1\}$ and $N_l=(N_{l-1}+2p_l-V_l+s_l)/s_l$ is the length and width of $\mathbf{Z}_l$.  The main benefit of this step is that only a subset of the feature maps are processed with the sliding filter at a time, which reduces the number of weights needed and hence improves computational performance.

\vspace{3pt}
\subsubsection{Activation Layer}
Activation functions perform nonlinear transformations to increase feature separation, which helps the network to learn complex patterns and solve nontrivial problems. The pre-trained transfer learning models considered in this paper use the rectified linear unit (ReLU) activation function. This activation function provides faster learning due to its constant gradient, which is crucial in deep learning architectures. Without loss of generality, suppose layer $l$ is a ReLU activation layer. Let $\mathbf{Z}_{l-1}$ be the 3D feature map input from the previous layer. The output feature map after the activation layer is given by:

\begin{equation}
    \label{eq:relu}
    \mathbf{Z}_l[i][j][k]=\max\big(0,\;\mathbf{Z}_{l-1}[i][j][k]\big)
\end{equation}
The output feature map  has the same dimensions as those of the input since activation is performed on each element. 

\vspace{3pt}
\subsubsection{Pooling Layer}
The pooling layer performs down-sampling by dividing the input  feature maps from the previous layer into (non-overlapping) rectangular regions and computing either the maximum or average of each region. This layer does not perform any learning; however, its purpose is to reduce the numbers of parameters and prevent overfitting. In the three transfer learning models considered, max pooling is often performed after the convolution and activation layers. Without loss of generality, suppose layer $l$ is a max pooling layer. Given a 3D feature map $\mathbf{Z}_{l-1}$ of size $N_{l-1}\times N_{l-1}\times C_{l-1}$ from the previous layer, a square pooling region of size $Q_l\times Q_l$, and a stride parameter $s_l\in\mathbb{Z}^+$, the output feature map is computed as:
\begin{equation}
    \label{eq:pooling}
    \mathbf{Z}_l[i][j][k]=\max_{x,y\in\{0,\dots,Q_l-1\}}\mathbf{Z}_{l-1}[s_l i+x][s_l j+y][k]
\end{equation}
where $i,j\in\{0,\dots,N_l-1\}$, $k\in\{0,\dots,C_{l-1}-1\}$, and $N_l=(N_{l-1}-Q_l+s_l)/s_l$ is the length and width of $\mathbf{Z}_l$.
{\blue}

\vspace{3pt}
\subsubsection{Flattening Layer}
Once feature extraction has been completed for each channel, the feature maps from each channel need to be reorganized and combined into a single flattened feature vector, denoted as the Multi-FM, for processing by the Tree Topology Classifier and the fully-connected networks therein. Suppose there are $L$ layers during the multi-channel feature extraction. Let $\mathbf{Z}_L^1$ and $\mathbf{Z}_L^2$ be the feature map outputs of the left and right channels, respectively, both with dimensions $N_L\times N_L \times C_L$. Each of these 3D feature maps are reshaped into 1D feature vectors, each with $N_L^2\cdot C_L$ elements. Then, the Multi-FM feature vector is created by concatenating the 1D feature vectors from each channel, which contains all $2\cdot N_L^2\cdot C_L$ elements from both $\mathbf{Z}_L^1$ and $\mathbf{Z}_L^2$.

\subsection{Classification}
DARE consists of hierarchical arranged tree-structured fully-connected neural network classifiers which are custom trained for systematically identifying individual classes, as shown in Figure~\ref{fig:training progress}. Since CADDY dataset has a large number of classes, a single neural network classifier may not yield satisfactory individual class recognition performances although it may still provide good overall average performance. The classification tree is constructed as follows. Each node of the tree represents a neural network classifier which is trained to separate certain groups of classes. The branches correspond to the outputs of the classifiers and the final decisions of individual classes are predicted at the leaf nodes. 

In the classification tree, the class groups separated at each node are formed based on similar categories (e.g., number of hands, hand orientation, number of fingers, etc.) The root node at level $0$, called RootNet, is trained to classify the three major categories: gestures, poses and true negatives (i.e., images with no, improper or wrong gestures). Subsequently, the two classifiers at level $1$, GNet10 and PNet11, further classify the gestures and poses identified from  RootNet, respectively. It is noted that the back and front faces of the hands with gloves have different features with the white square shape on the back and white circle on the front. Moreover, the gestures are made with either one or two hands. Thus, GNet10 is constructed as a quinary classifier to classify the following five gesture types: one hand (back face), one hand (front face), two hands closed (back face), two hands open (back face), two hands open (front face). On the other hand, PNet11 is constructed as a ternary classifier to separate the three different diver poses: turning horizontally, turning vertically and free swim. Next, at level $2$, the one hand gestures are further classified using the finger/thumb pointing direction and the number of fingers. Thus, GNet20 is constructed as a ternary classifier to classify the one hand (back face) gestures into three  categories: thumb pointing up/no finger, finger pointing down/front, and finger(s) pointing up. Similarly, GNet21 is constructed as a quaternary classifier to classify the one hand (front face) gestures into four categories: thumb pointing down, four/five fingers, three fingers/two fingers and a thumb, and one/two fingers.

Finally, at level $3$, six binary classifiers are constructed: GNet30, GNet31, GNet32, GNet33, GNet34 and GNet35, which classify no finger vs thumb; finger pointing downward vs finger pointing forward; two fingers pointing up vs one finger pointing up;  four vs five fingers pointing up; three fingers vs two fingers and a thumb pointing up; and two fingers vs one finger pointing up, respectively.

As previously mentioned, each node of the tree is a fully-connected feed-forward neural network classifier. The input layer to each of these classifiers is the Multi-FM. The architecture of the hidden layers of each of these classifiers is the same as the fully-connected hidden layers of the pre-trained transfer learning model that the multi-channel feature extraction stages are based on. For example, AlexNet has two identical fully-connected hidden layers. Each hidden layer has $4096$ neurons that are connected to each input from the previous layer, a ReLU activation layer, and a $50\%$ dropout layer. Thus, if the convolution layers of AlexNet are used for multi-channel feature extraction, then these hidden layers are used in each node classifier of the tree. These hidden layers are then followed by a fully-connected output layer that has the same number of neurons as number of classes or class groups separated at that node, and a softmax layer to calculate the probability of a sample belonging to a class. The weights of the fully-connected network are retrained using the cross-entropy loss function at the output.

The tree-structured classifier simplifies the 20-class underwater diver action recognition problem by constructing several focused neural network classifiers arranged in a hierarchical manner. It not only boosts the individual class recognition performance but also improves the overall performance, while producing a robust and reliable decision in real-time.

\subsection{Performance Measures}

In this paper, we used several measures to evaluate the performance of DARE against other existing deep learning architectures. For each individual diver action, we compute the micro F1 score and balanced individual class accuracy. Then, we compute the macro F1 score and the overall correct classification rate (CCR). Each of these measures are defined in terms of the number of true positives, false negatives, true negatives, and false positives for each diver action.
\vspace{3pt}

For each class $i\in\{1,\dots,n\}$, where $n$ is the number of classes, let $TP_i$ be the number of true positives, which is the number of samples of class $i$ that are correctly classified. Let $FN_i$ be the number of false negatives, which is the number of samples of class $i$ that are misclassified. It should be noted that the total number of samples belonging to class $i$ is  $TP_i+FN_i$. Also let $TN_i$ be the number of true negatives, which is the number of samples belonging to any class $j\neq i$ that are not classified as class $i$. Finally, let $FP_i$ be the number of false positives, which is the number of samples belonging to any class $j\neq i$ that are misclassified as class $i$.

\vspace{3pt}
\textit{F1 Score}: The F1 Score is the harmonic mean of precision and recall. A high precision for diver action $i$ suggests that there are few false positives relative to true positives (i.e., other diver actions are not misclassified as action $i$), whereas a high recall suggests that there are few false negatives relative to true positives (i.e., diver action $i$ is not misclassified as another action). The precision and recall for class $i$ are computed as:
 
\begin{equation}
\label{eq:precision}
Precision_i = \frac{TP_i}{TP_i+FP_i},  \ i = 1,\dots,n,
\end{equation}
\begin{equation}
\label{eq:recall}
Recall_i = \frac{TP_i}{TP_i+FN_i}, \ i = 1,\dots,n.
\end{equation}
Then, the individual class (micro) F1 score is computed as:
\begin{equation}
\label{eq:microF1}
F\textit{1}_{i} = 2\cdot \frac{Precision_i\times Recall_i}{ Precision_i + Recall_i}, \ i = 1,\dots,n
\end{equation}
and the average (macro) F1 score is computed as:
\begin{equation}
    \label{eq:macroF1}
    \overline{F\textit{1}} = \frac{1}{n}\sum_{i=1}^{n}F\textit{1}_{i}.
\end{equation}

\textit{Balanced Individual Class Accuracy}: The balanced individual class accuracy (BACC) is the mean of the true positive rate and true negative rate. This measure is useful in inherently unbalanced one-verses-the-rest class performance comparisons, as other accuracy measures do not normalize the true positive and true negative predictions. The individual class true positive rate $TPR_i$ is the same as the recall in Eq.~(\ref{eq:recall}), and the individual class true negative rate is computed as:
\begin{equation}
    \label{eq:tnr}
    TNR_i=\frac{TN_i}{TN_i+FP_i} \ i = 1,\dots,n.
\end{equation}
Then, the balanced individual class accuracy is computed as:
\begin{equation}
    \label{eq:balancedacc}
    BACC_i=\frac{TPR_i+TNR_i}{2}, \ i = 1,\dots,n.
\end{equation}

\textit{Correct Classification Rate}: The CCR is the overall accuracy of the classifier, which is computed as:
\begin{equation}
    \label{eq:ccr}
    CCR = \frac{\sum_{i=1}^{n} TP_i}{\sum_{i=1}^n TP_i+FN_i}\times 100\%
\end{equation}

\begin{figure*}[t]
    \centering
   \subfloat[Balanced accuracy of ResNet-based networks ]{\includegraphics[width=0.335\textwidth]{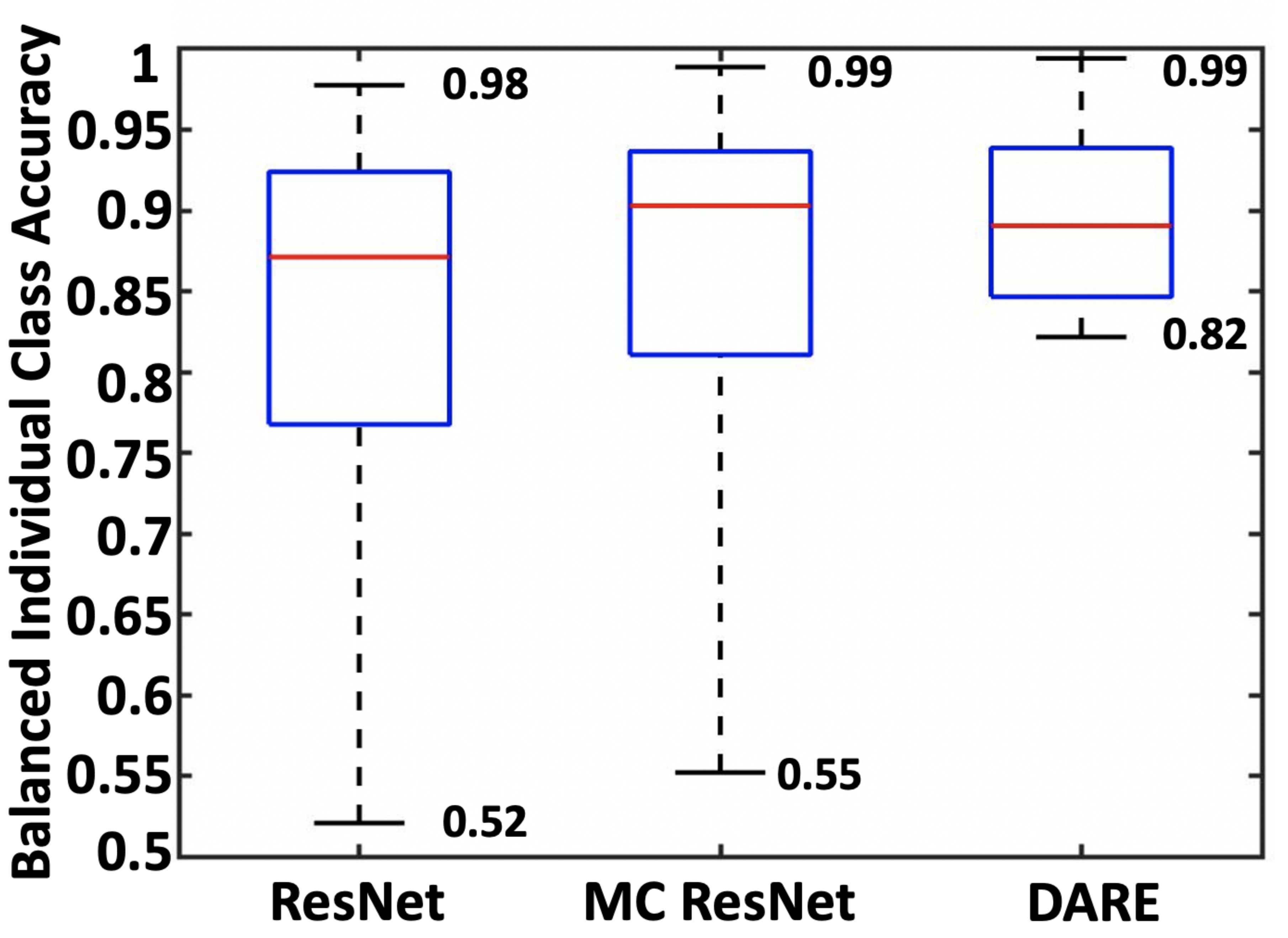}\label{fig:boxba_res}}   
    \centering   
    \subfloat[Balanced accuracy  of AlexNet-based networks]{\includegraphics[width=0.335\textwidth]{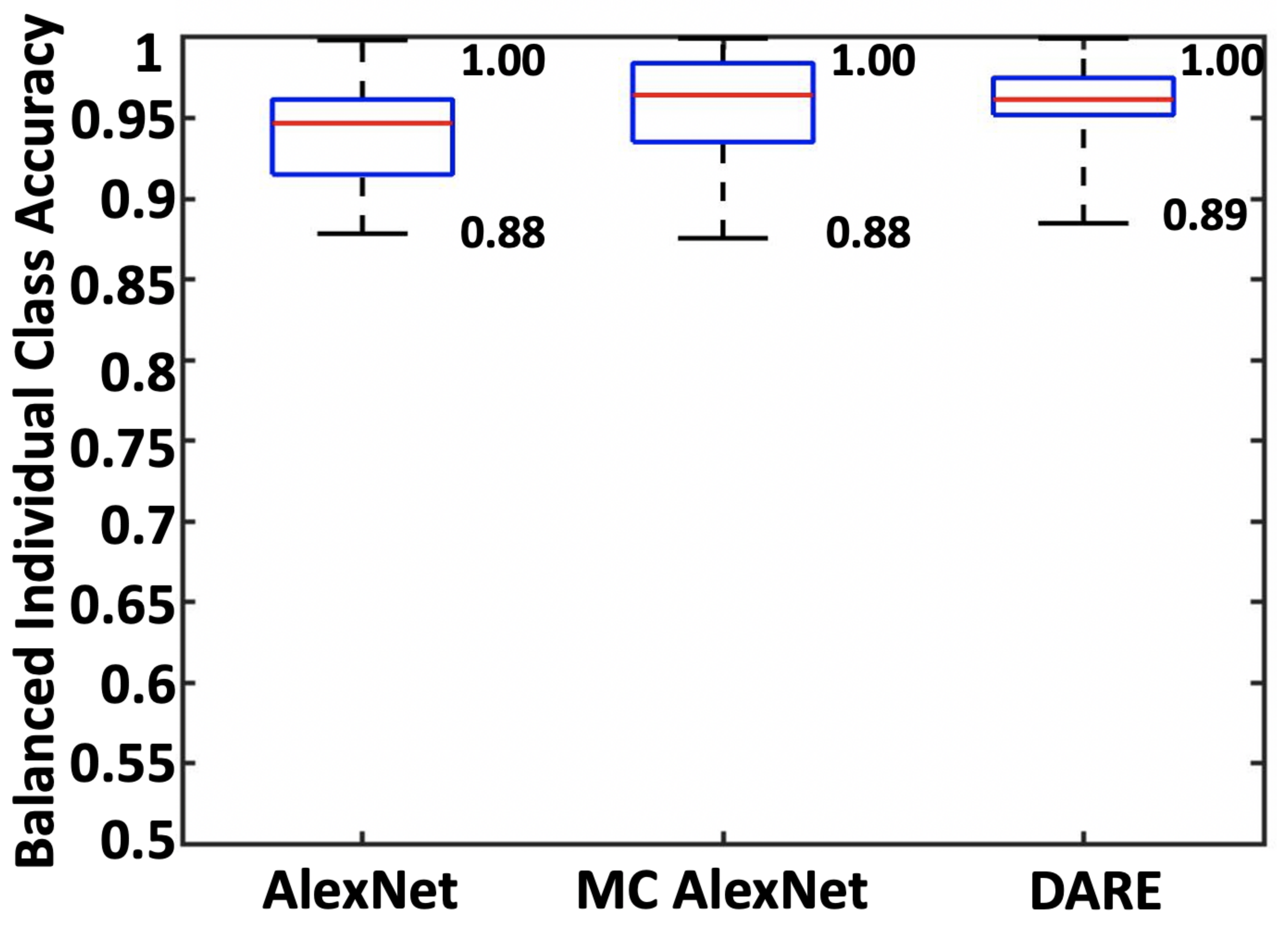}\label{fig:boxba_alex}}
    \centering
    \subfloat[Balanced accuracy  of VggNet-based networks]{\includegraphics[width=0.335\textwidth]{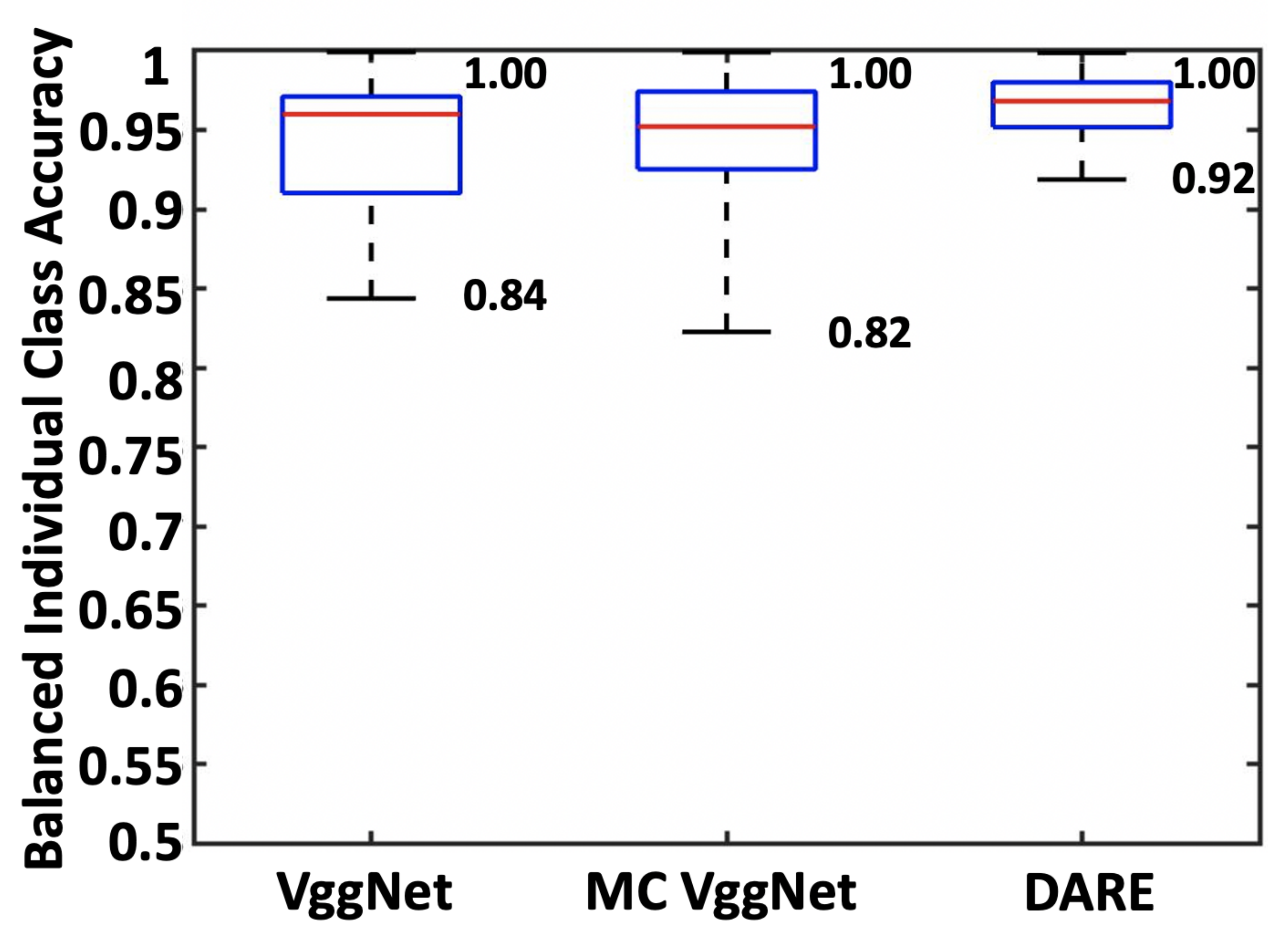}\label{fig:boxba_vgg}}
    \\
    \centering
   \subfloat[F1 scores of ResNet-based networks ]{\includegraphics[width=0.335\textwidth]{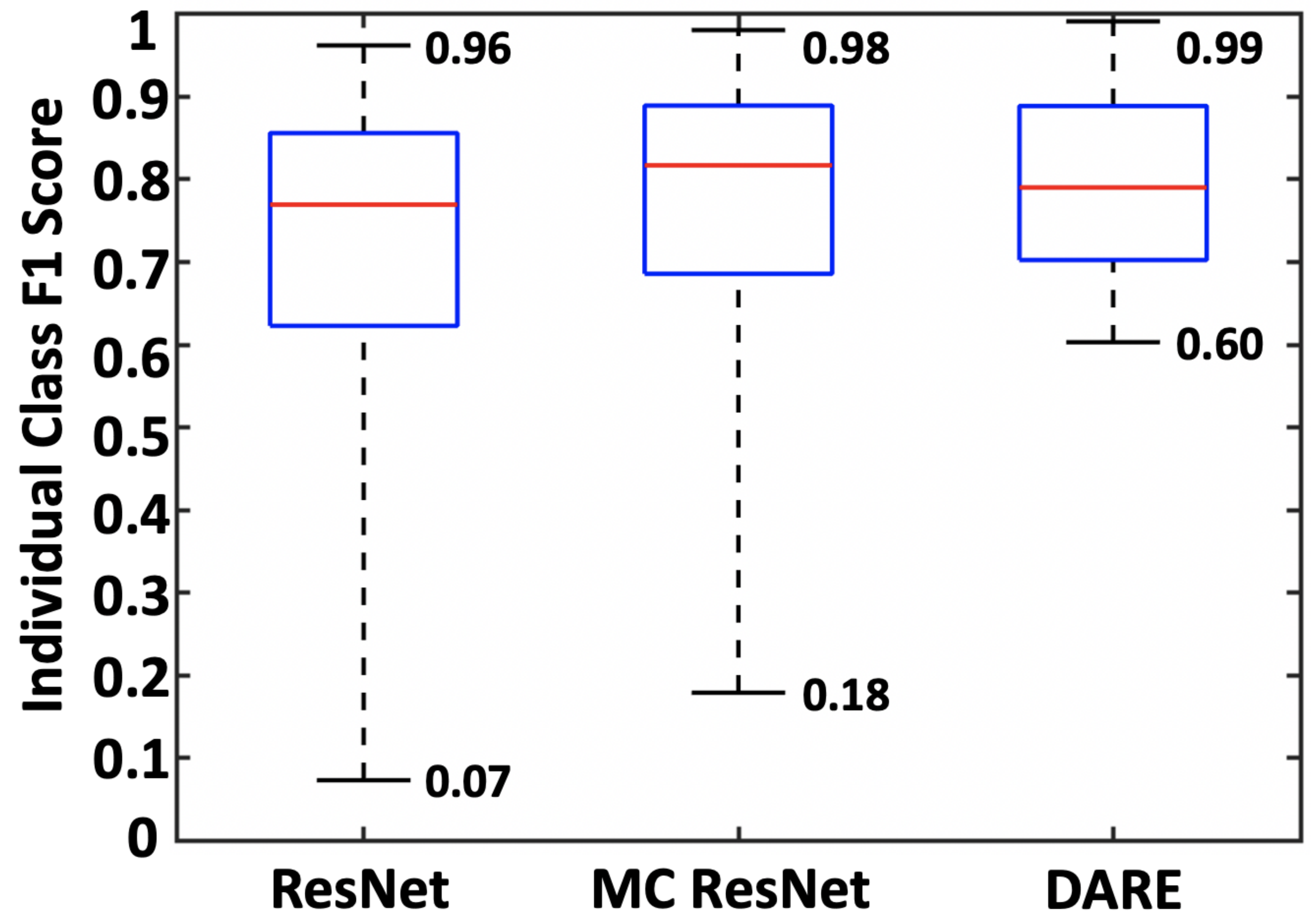}\label{fig:boxfscore_res}}   
    \centering   
    \subfloat[F1 scores of AlexNet-based networks]{\includegraphics[width=0.335\textwidth]{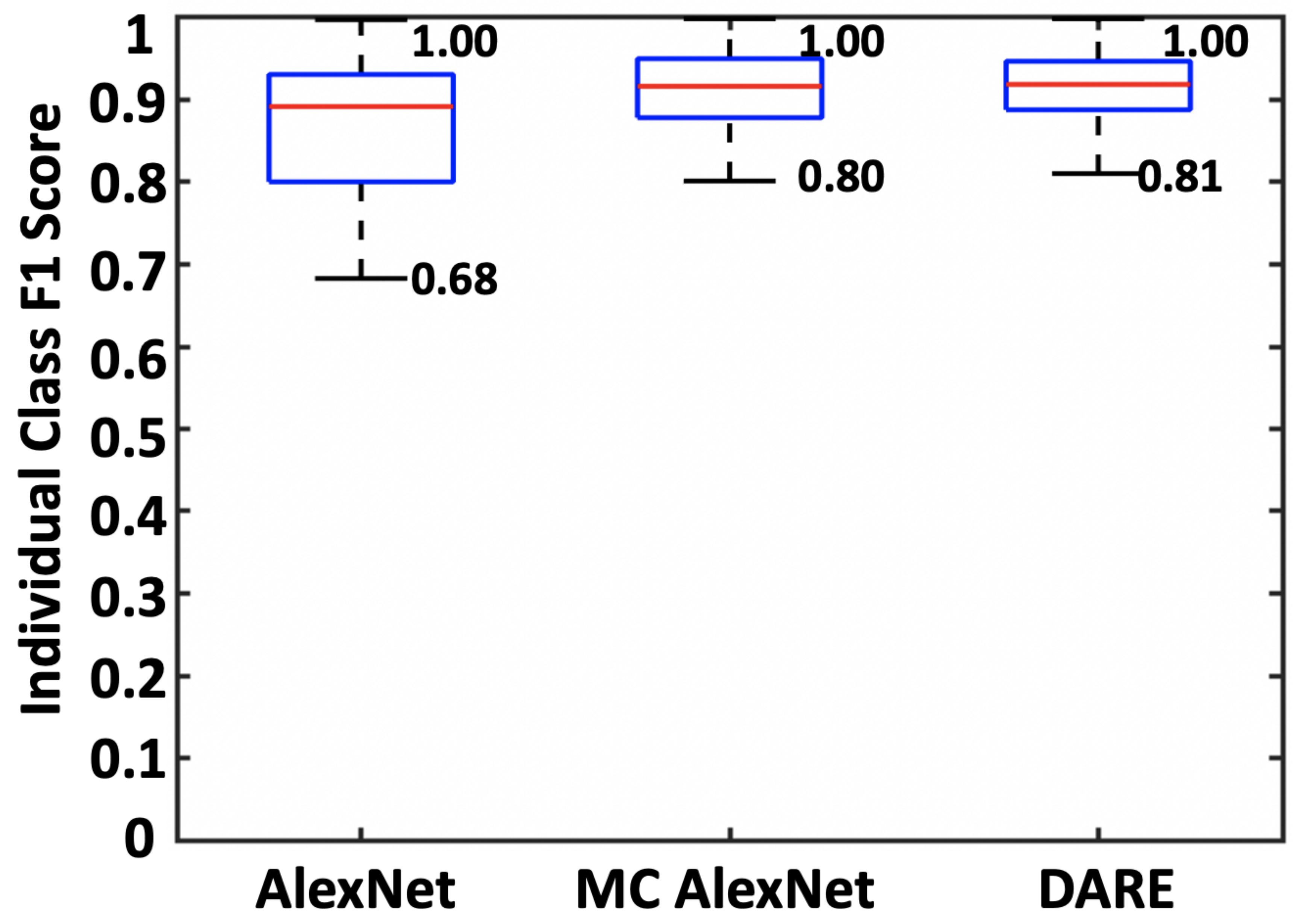}\label{fig:boxfscore_alex}}
    \centering
    \subfloat[F1 scores of VggNet-based networks]{\includegraphics[width=0.335\textwidth]{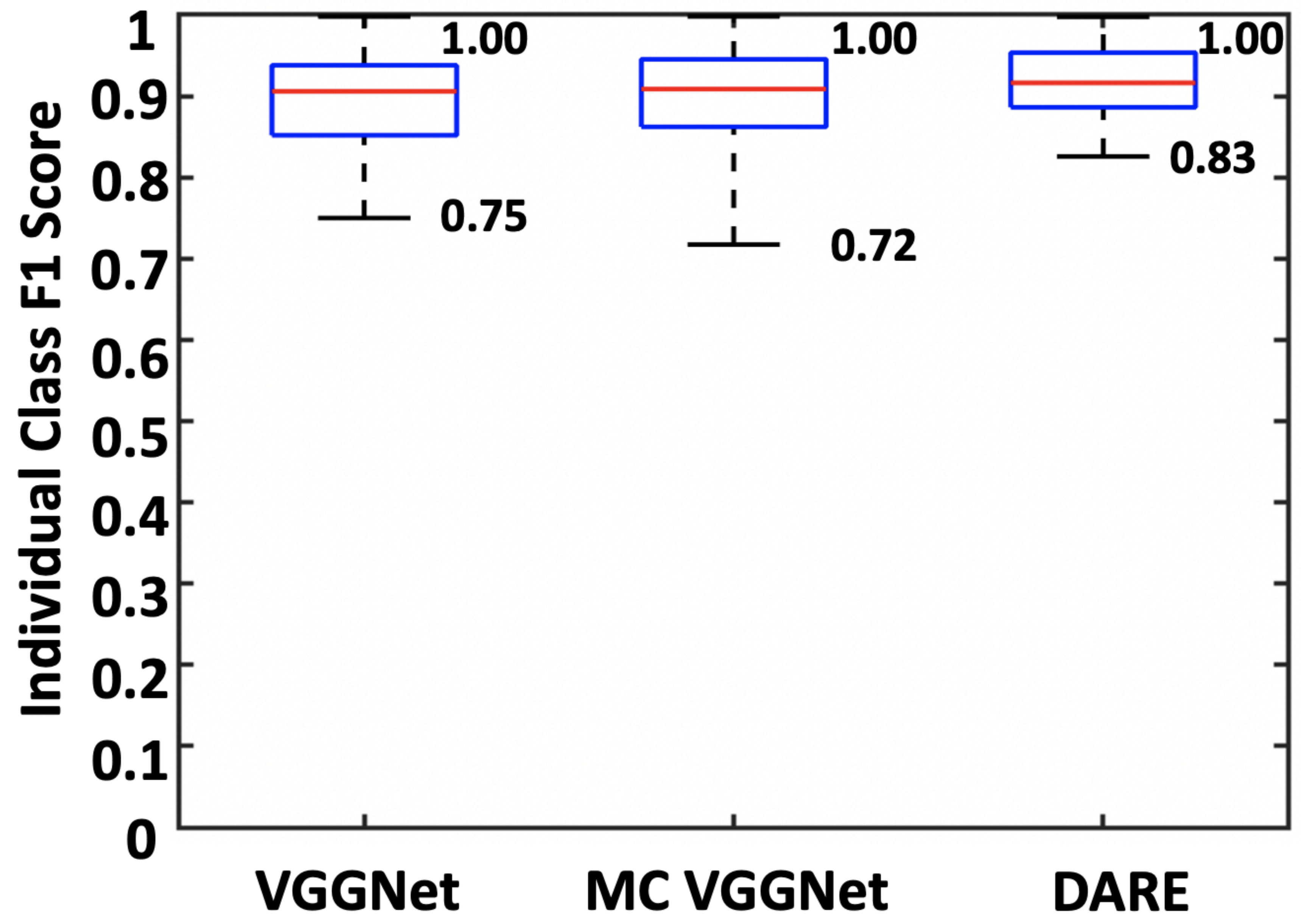}\label{fig:boxfscore_vgg}}
\caption{Box plots of the balanced individual class accuracy and (micro) F1 score for different learning networks.}
\label{fig:boxplots} \vspace{-10pt}
\end{figure*}

\section{Results and Discussion}
\label{sec:result}
This section presents the comparative evaluation results of the performance of DARE with other deep transfer learning methods. In this paper, three base networks are used: ResNet18, AlexNet, and VggNet16. Furthermore, each of these base networks are augmented in the bi-channel framework to analyse stereo image pairs. Subsequently, DARE is built from each of these bi-channel networks by replacing the single neural network classifier with a tree-topology classifier to boost the individual class recognition performance. Thus, the performance of DARE networks with three different underlying transfer learning models are compared with the corresponding base networks and their bi-channel augmented networks. Classification was performed on the $16$ hand gesture types, $3$ diver pose types, and $1$ miscellaneous class of  no-gesture/no-pose images. In order to utilize the transfer learning approaches above, the CADDY dataset images were downsampled and resized to the specified input image sizes of each network architecture. 
The data analysis was performed using the MATLAB Deep Learning Toolbox on a Windows 10 computer with an Intel Core i7 7700 processor and 32GB of RAM. For DARE, in each of the sub-classifiers in the tree, the initial learning rate is set to $0.001$ with stochastic gradient descent with momentum optimizer. The performance results are obtained  via the k-fold cross validation.

Figure \ref{fig:boxplots} shows the variations of individual class performance measures using the box plots. While Figures \ref{fig:boxba_res}-\ref{fig:boxba_vgg} show the box plots of balanced individual class accuracies for different networks, Figures \ref{fig:boxfscore_res}-\ref{fig:boxfscore_vgg} show the boxplots of individual class F1-scores for different networks. Each boxplot shows a red line (the median), two black lines (the minimum and maximum values), and a blue box with the lower and upper boundaries representing the $25^{th}$ and $75^{th}$ percentile values, respectively. Figure \ref{fig:boxba_res} shows the box plots of balanced individual class accuracy of the ResNet-based networks. As seen, the regular ResNet has a huge variation in the accuracy values with the maximum accuracy of $0.98$ and the minimum accuracy of $0.52$, which will give poor underwater performance. The blue box also shows high variation. On the other hand, MC ResNet not only boosts the maximum and minimum accuracy values to $0.99$ and $0.55$, respectively, it also shrinks the variation between the $25^{th}$ and $75^{th}$ percentile values. On the other hand, DARE significantly improves the minimum accuracy value to $0.82$, thus verifying the utility of the tree-topology classifier. It also shrinks the interval between the $25^{th}$ and $75^{th}$ percentile values, where these values are equal to  $0.84$ and $0.94$, respectively, which are better than those of the regular ResNet and MC ResNet. 

\begin{table*}[t]
\normalsize
\caption{Comparison of the overall average performance measures of different learning networks.}\vspace{-6pt}
\begin{center}
\begin{tabular}{c|ccc|ccc|ccc}
\hline

\multirow{2}{*}{Metrics} & \multicolumn{3}{c|}{ResNet} &\multicolumn{3}{c|}{AlexNet} &\multicolumn{3}{c}{VggNet}\\
\cline{2-10}
 &\tabincell{c}{Regular} & \tabincell{c}{MC}& \tabincell{c}{DARE} &\tabincell{c}{Regular} &\tabincell{c}{MC} & \tabincell{c}{DARE} &\tabincell{c}{Regular} & \tabincell{c}{MC}& \tabincell{c}{DARE}\\

\hlx{vhvv}

\tabincell{l}{CCR (\%)} & \tabincell{l}{86.03}&\tabincell{c}{88.80}& \tabincell{c}{89.47}& \tabincell{c}{94.21} & \tabincell{l}{95.88} & \tabincell{c}{95.93}& \tabincell{c}{95.20}& \tabincell{l}{95.39}& \tabincell{c}{95.87}\\
\hlx{vv}

\tabincell{l}{Macro F1 score}  & \tabincell{c}{0.728}& \tabincell{l}{0.782}& \tabincell{c}{0.799}& \tabincell{c}{0.875} & \tabincell{l}{0.919} & \tabincell{c}{0.920} & \tabincell{c}{0.899}& \tabincell{l}{0.901}& \tabincell{c}{0.921}\\
\hlx{vv}

\tabincell{l}{Training \\ Time (hrs)} &\tabincell{c}{1.46} &\tabincell{l}{1.72} & \tabincell{c}{3.65}& \tabincell{c}{2.08} & \tabincell{l}{3.92} & \tabincell{c}{13.80} & \tabincell{c}{7.82} &\tabincell{l}{13.80}&\tabincell{c}{17.19}\\
\hlx{vv}

\tabincell{l}{Testing \\ Time (ms)} & \tabincell{c}{34.59}& \tabincell{l}{69.24}& \tabincell{c}{72.65}&\tabincell{c}{16.69} & \tabincell{l}{33.33} & \tabincell{c}{34.20} & \tabincell{c}{266.75}& \tabincell{l}{ 533.09} & \tabincell{c}{534.49}\\
\hline

\end{tabular}
\label{table_result}\vspace{-12pt}
\end{center}
\end{table*}

Similarly, Figure \ref{fig:boxba_alex} shows the box plots of balanced individual class accuracy of the AlexNet-based learning networks. As seen, DARE outperforms the regular AlexNet and MC AlexNet with the minimum accuracy of $0.89$. DARE also achieves the shortest range of the $25^{th}$ and $75^{th}$ percentile values. Finally, Figure \ref{fig:boxba_vgg} shows the box plots of balanced individual class accuracy of the VggNet-based learning networks. As seen, DARE significantly boosts the minimum accuracy to $0.92$, while shrinking the $25^{th}$ and $75^{th}$ percentile range, which indicates significantly better and reliable performance as compared to the regular VggNet and MC VggNet. 

Figures \ref{fig:boxfscore_res}-\ref{fig:boxfscore_vgg} show similar trends in the boxplots of individual class F1 scores for different networks, where for each base network DARE supersedes the performance of the other networks by boosting the minimum individual class F1 score and shrinking the range of the $25^{th}$ and $75^{th}$ percentile values. In particular, DARE with VggNet-based training model achieves the best performance with the minimum individual class F1 score of $0.83$.

Table \ref{table_result} presents the overall CCR, overall F1 score, and the training and testing times of all networks. DARE gives the highest overall CCR and F1 score among all base networks. Although DARE requires slightly more training times due to training of several sub-classifiers in the tree; its testing times are fairly small and are suitable for real-time implementation. 

In summary, DARE with  VggNet-based training model reveals the best performance in terms of both the individual class and  overall performance measures. The high accuracies, smallest range of the $25^{th}$ and $75^{th}$ percentile values and high F1-scores obtained by DARE indicate that it delivers robust and reliable performance in the presence of various environmental, driver and sensor uncertainties in the CADDY dataset. Although DARE with VggNet gives the best classification performance, the prediction time for one stereo pair is $0.53$ seconds. While this is suitable for real-time prediction in static or slowly changing environments, it might not be adequate in rapidly changing environments with sporadic diver movements. On the other hand, DARE with AlexNet can make predictions in only $0.034$ seconds, while offering diver action recognition performance close to DARE with VggNet. As such, DARE with AlexNet is able to process all incoming frames from the stereo camera system on the AUV in order to make robust and reliable predictions in real-time under dynamic underwater operating conditions.

\section{Conclusion and Future Work}
\label{sec:conclusion}
This paper addresses the diver gesture and pose  recognition problem for robust, efficient and reliable underwater communication between the diver and AUV in uncertain underwater environments. In this regard, the paper developed a diver action recognition system, called DARE, which is trained using the CADDY dataset collected in real underwater environments. DARE is built upon a multi-channel deep transfer learning architecture with a tree-topology neural network classifier. The multi-channel deep transfer learning CNN facilitates feature extraction and fusion from stereo-pairs of AUV camera images. Furthermore, the tree structured neural network based classifier boosts the individual class recognition performance for reliable operation. The results show that DARE delivers better performance in both individual class and overall classification measures. In addition, DARE is computationally efficient and delivers fast decisions, making it suitable for real-time underwater implementation.

For future work, there are several areas of research for this problem. First, to further enhance the classification performance, a customized multi-channel CNN can be designed to extract specific and useful features. Second, more data can be collected to train the classifier with more different diver actions under diverse real-life underwater conditions. Finally, a smart information-theoretic approach to efficiently and automatically create the tree topology classifier can be developed so that DARE can scale better to arbitrarily large diver action recognition problems. 

\bibliographystyle{ieeetr}
\bibliography{DARE}
\end{document}